\documentclass[10pt,twocolumn,letterpaper]{article}

\usepackage[pagenumbers]{cvpr} 

\usepackage{xspace}

\usepackage{booktabs}
\usepackage{pifont}
\newcommand{\cmark}{\color{ForestGreen}\ding{51}}
\newcommand{\xmark}{\color{red}\ding{55}}

\usepackage{multirow}
\usepackage{multicol}

\newcommand{\pamuk}{\textsc{PaMuk-Bench}\xspace}
\newcommand{\std}[1]{\textsubscript{$\pm$#1}}

\definecolor{cvprblue}{rgb}{0.21,0.49,0.74}
\usepackage[pagebackref,breaklinks,colorlinks,allcolors=cvprblue]{hyperref}

\def\confName{CVPR}
\def\confYear{2027}

\title{\vspace{-0.0cm}\textsc{PaMuk-Bench}: A Parallel Multilingual Multitask Benchmark for Vision-Language Tasks\vspace{-0.25cm}}

\author{
    Silviu-Florin Gheorghe\textsuperscript{\rm 1},
    Radu-Stefan Popa\textsuperscript{\rm 1},
    Adriana Valentina Costache\textsuperscript{\rm 1}, 
    Adrian-Gabriel Chifu\textsuperscript{\rm 2},\\
    Alessia Amelio\textsuperscript{\rm 3}, 
    Paul Irofti\textsuperscript{\rm 1},
    Anders Skaarup Johansen\textsuperscript{\rm 4},
    Lallouani Bouchakour\textsuperscript{\rm 5},\\
    Mohamed Lichouri\textsuperscript{\rm 6},
    Khaled Lounnas\textsuperscript{\rm 5},
    Neelu Madan\textsuperscript{\rm 4,7},
    Yashvir Malik\textsuperscript{\rm 4,8},
    Radu Tudor Ionescu\textsuperscript{\rm 1,$\diamond$}\\
    \textsuperscript{\rm 1}Univ.~of Bucharest\quad
    \textsuperscript{\rm 2}Aix-Marseille Univ.\quad
    \textsuperscript{\rm 3}Univ.~of Chieti-Pescara\quad
    \textsuperscript{\rm 4}Aalborg Univ.\\
    \textsuperscript{\rm 5}CRSTDLA Research Center\quad
    \textsuperscript{\rm 6}Blida 1 Univ.\quad
    \textsuperscript{\rm 7}Poineer Centre AI\quad
    \textsuperscript{\rm 8}Central Univ.~of Haryana\\
\textsuperscript{$\diamond$}Corresp.~author: raducu.ionescu@gmail.com
\vspace{-0.6cm}}

\begin{document}
\twocolumn[{%
\renewcommand\twocolumn[1][]{#1}%
\maketitle
\begin{center}
    \centering
    \captionsetup{type=figure}
    \includegraphics[width=0.94\textwidth]{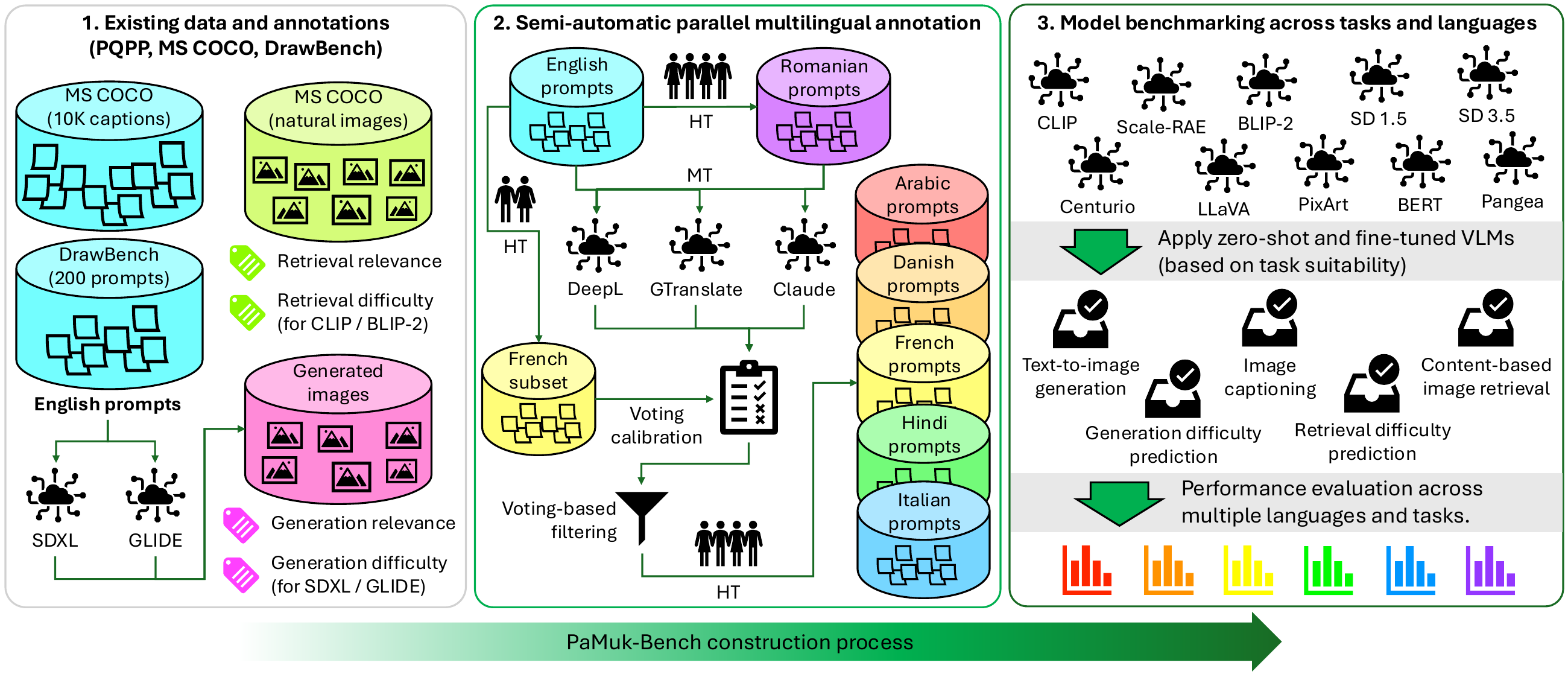}
    \vspace{-0.35cm}
    \captionof{figure}{Overview of \pamuk. (1) Our benchmark harnesses English prompts from PQPP \cite{poesina_CVPR_2025}, natural images from MS COCO \cite{lin_LNSC_2014, chen_arxiv_2015}, and synthetic images generated by SDXL \cite{Podell-ICLR-2024} and GLIDE \cite{Nichol-ICML-2021}, along with generation/retrieval relevance judgments, as well as prompt difficulty annotations for both generation and retrieval. (2) We translate every source prompt via a semi-automatic process, leveraging voting based on machine translation (MT) tools (DeepL \cite{deepl}, Google Translate \cite{googletranslate}, Claude sonnet 4.6 \cite{claude}) to reduce human translation (HT) effort. (3) Multiple vision-language models (VLMs) are evaluated on \pamuk (according to task suitability), considering both zero-shot and fine-tuning setups. Models are evaluated across five tasks and seven languages. Best viewed in color.}
    \label{fig:teaser}
\end{center}%
}]

\begin{abstract}
While vision-language models (VLMs) currently obtain notable results for highly-studied languages, \eg English, they exhibit significant performance drops in certain low-resource languages, especially due to the lack of annotated datasets. To this end, we propose \pamuk, a novel \textbf{pa}rallel \textbf{mu}ltilingual multitas\textbf{k} benchmark for vision-language tasks. Our benchmark includes a set of over 10K curated English prompts that are translated into six languages: Arabic, Danish, French, Hindi, Italian, Romanian. Parallel translations, which are obtained via a semi-automatic process, along with accompanying generated and retrieved images enable the comparative development and evaluation of VLMs across a wide range of tasks: text-to-image generation, image captioning, content-based image retrieval, image generation difficulty estimation, image retrieval difficulty estimation. We empirically show that zero-shot English and multilingual pre-trained VLMs compare consistently worse than their fine-tuned counterparts, confirming that \pamuk is a valuable resource that can drastically reduce the language gap across a variety of VLM architectures and tasks. \pamuk is available at \small{\url{https://github.com/ghesil/pamuk_bench}}.
\end{abstract}

\vspace{-0.3cm}
\section{Introduction}
\label{sec:intro}
\vspace{-0.2cm}

Vision-language understanding (VLU) has seen tremendous progress in recent years due to the development of foundation vision-language models (VLMs) \cite{laurenccon_nips_2024, liu2023visual,zhang_tpami_2024}, harnessing the capabilities of large language models (LLMs) \cite{Touvron-A-2023,openai2023gpt4,jiang2023mistral7b} coupled with the effective infusion of visual information. However, the scarcity of multilingual resources for VLU tasks \cite{thapliyal2022crossmodal,srinivasan2021wit,elliott2016multi30k}, especially across low-resource languages \cite{lupacscu2026large}, hindered the developmental parity across different languages, leading to a significant performance gap when switching from English to a low-resource language. Vision-language datasets generally address specific tasks, \eg text-to-image generation \cite{kirstain2023pick,xu2023imagereward,yu2022scaling} or image captioning \cite{young-etal-2014-image, chen_arxiv_2015}, in a mono-lingual (English) setup. Most multilingual benchmarks also comprise a limited set of tasks, \eg image retrieval \cite{srinivasan2021wit,bugliarello-etal-2022-iglue, aggarwal2021towards} or image captioning \cite{2024-lovenia, thapliyal2022crossmodal,elliott2016multi30k}, but the collections are usually heterogeneous across languages. To the best of our knowledge, there are no parallel multilingual benchmarks for vision-language understanding that cover multiple diverse tasks, ranging from image generation and captioning to prompt/query difficulty prediction \cite{poesina_CVPR_2025}.

To assess and reduce the multilingual parity gap, we introduce \pamuk, a novel \textbf{pa}rallel \textbf{mu}ltilingual multitas\textbf{k} benchmark for vision-language understanding that covers seven languages (see \cref{fig:teaser}). More specifically, our benchmark includes a set of 10,200  English prompts that are translated into six languages: Arabic, Danish, French, Hindi, Italian, Romanian. We employ a semi-automatic translation process, where 49,505 sentences are manually translated by native speakers of the target languages, and 11,695 are automatically translated, upon finding an agreement among three machine translation (MT) models. Parallel translations along with accompanying generated and natural images enable the comparative development and evaluation of VLMs across a wide range of VLU tasks. In particular, \pamuk covers text-to-image generation, image captioning, content-based image retrieval, image generation difficulty prediction, and image retrieval difficulty prediction. For all five tasks, \pamuk comes with official training, validation and test splits, enabling fair comparisons among future studies without requiring the evaluation of previous models on different data splits. The parallel prompts coupled with the fixed data splits further enable direct intra- and inter-language comparisons. \pamuk includes generated images by SDXL \cite{Podell-ICLR-2024} and GLIDE \cite{Nichol-ICML-2021} using the original English prompts, as well as natural images from MS COCO \cite{lin_LNSC_2014}. For both generated and natural images, we harness the relevance judgments from the public PQPP benchmark \cite{poesina_CVPR_2025}.

On the five vision-language understanding tasks, we evaluate several state-of-the-art VLMs, which are specifically designed for each target task. Our results uncover some interesting findings. First, zero-shot models perform significantly worse on non-English languages than on English, indicating a severe performance drop when English text is replaced with an equivalent translation. This observation generally holds for both English and multilingual pre-trained models. Second, the performance gap can be recovered, in most part, by simply fine-tuning the VLMs on each of the target tasks, regardless of model architecture or pre-training data. Put together, these findings indicate that \pamuk is a valuable resource that can both effectively assess and drastically reduce the language gap across a variety of VLM architectures and tasks.

In summary, our contribution is threefold:
\begin{itemize}
    \item We introduce \pamuk, a novel \textbf{pa}rallel \textbf{mu}ltilingual multitas\textbf{k} benchmark covering seven languages, which can be used to both assess and reduce the inter-language performance parity gap of vision-language models across a variety of languages.
    \item We employ a semi-automatic translation process to obtain 61,200 translated prompts, out of which 49,505 are manually corrected or translated by proficient speakers of Arabic, Danish, French, Hindi, Italian, and Romanian. 
    \item We evaluate several state-of-the-art VLMs, in both zero-shot and fine-tuning setups, across five VLU tasks, (i) exposing the English vs.~non-English performance parity gap, and (ii) showing that it can be almost entirely recovered via fine-tuning.
\end{itemize}

    \begin{table*}[t]
    \centering
    \caption{Comparison between PaMuk-Bench and related benchmarks. There are two groups of benchmarks: monolingual (English) and multilingual. \emph{Parallel} denotes identical images and annotations shared across languages. Task coverage considers only task overlaps with \pamuk. For benchmarks that include pure NLP or audio tasks, the numbers are reported for the subset of vision-language tasks.}
    \label{tab:related-datasets}
    \vspace{-0.35cm}
    \renewcommand{\arraystretch}{1.1}
    \fontsize{7.3}{7.5}\selectfont{
    \setlength{\tabcolsep}{2.35pt} 
    \begin{tabular}{|l|cccc|ccccc|cc|}
    \hline
     \multirow{3.8}{*}{Benchmark} & \multirow{3.6}{*}{\rotatebox[origin=c]{40}{\#Languages}} & \multirow{3.6}{*}{\rotatebox[origin=c]{40}{Parallel}} & \multirow{3.6}{*}{\rotatebox[origin=c]{40}{\#Prompts}} & \multirow{3.6}{*}{\rotatebox[origin=c]{40}{\#Images}} & \multicolumn{5}{c|}{Vision-Language Tasks} & \multicolumn{2}{c|}{Image Types} \\
    \cline{6-10}
    \cline{11-12}
    & & & & &
    \rotatebox[origin=c]{20}{Im.~Generation} & \rotatebox[origin=c]{20}{Im.~Retrieval} & \rotatebox[origin=c]{20}{Im.~Captioning} & \rotatebox[origin=c]{20}{Gen.~Difficulty} & \rotatebox[origin=c]{20}{Ret.~Difficulty} & \rotatebox[origin=c]{20}{Natural} & \rotatebox[origin=c]{20}{Generated} \\
    \hline    
    MS COCO \cite{lin_LNSC_2014, chen_arxiv_2015} & 1 (En)  & n/a & 1M & 328,000 &  \xmark & \xmark & \cmark & \xmark & \xmark &  \cmark & \xmark \\
    Flick30k \cite{young-etal-2014-image} & 1 (En)  & n/a & 158,915 & 31,783 & \xmark & \xmark & \cmark & \xmark &  \xmark & \cmark & \xmark \\
    Visual Genome \cite{krishna2017visual} & 1 (En)  & n/a  & 5M & 108,077 & \xmark & \xmark & \cmark & \xmark & \xmark &  \cmark & \xmark \\
    PartiPrompts \cite{yu2022scaling} & 1 (En)  & n/a & 1,600 & 0 & \cmark & \xmark & \xmark & \xmark & \xmark &  \xmark & \xmark \\
    ImageReward \cite{xu2023imagereward} & 1 (En) & n/a & 8,878 & 62,200 & \cmark & \xmark & \xmark & \xmark & \xmark & \xmark & \cmark \\
    Pick-a-Pic \cite{kirstain2023pick} & 1 (En) & n/a & 35,000 & 500,000 & \cmark & \xmark & \xmark & \xmark & \xmark & \xmark  & \cmark \\
    iQPP \cite{poesina2023iqpp}      & 1 (En) & n/a & 2,940 & 37,726 & \xmark & \xmark & \xmark & \xmark & \cmark  & \cmark & \xmark \\    
    PQPP \cite{poesina_CVPR_2025}    & 1 (En) & n/a & 10,200 & 368,800 & \xmark & \xmark & \xmark & \cmark & \cmark & \cmark & \cmark \\
\hline
    SEACrowd \cite{2024-lovenia} & 4 & \xmark & 11,188 & 3,600 & \cmark & \xmark & \cmark & \xmark & \xmark & \cmark & \xmark \\
    
    WIT \cite{srinivasan2021wit} & 108 & \xmark & 37M & 11.5M & \xmark & \cmark & \xmark & \xmark & \xmark & \cmark & \xmark \\

    IGLUE \cite{bugliarello-etal-2022-iglue}    & 20 & \xmark & n/a & 700,000 & \xmark & \cmark & \xmark & \xmark & \xmark & \cmark & \xmark \\
    
    XM3600 \cite{thapliyal2022crossmodal}  & 36 & \xmark & 261,375 & 3,600 & \xmark & \xmark & \cmark & \xmark & \xmark & \cmark  & \xmark \\

    Multi30K \cite{elliott2016multi30k} & 2 & partial & 155,070 & 31,783 & \xmark & \xmark & \cmark & \xmark & \xmark & \cmark  & \xmark \\

    XTD10~\cite{aggarwal2021towards} & 11 & \cmark & 11,000 & 1,000 & \xmark & \cmark & \xmark & \xmark & \xmark & \cmark  & \xmark \\
    
    \hline
    \pamuk (ours)   & 7 & \cmark & 71,400 & 368,800 & \cmark & \cmark & \cmark & \cmark & \cmark & \cmark & \cmark\\
    \hline
    \end{tabular}
    }
     \vspace{-0.4cm}
    \end{table*}

\vspace{-0.1cm}
\section{Related Work}\label{sec:related}
\vspace{-0.2cm}

\noindent \textbf{Vision-language benchmarks.}
Evaluation of vision-language models has traditionally relied on English language datasets \cite{lin_LNSC_2014, chen_arxiv_2015, young-etal-2014-image, krishna2017visual, yu2022scaling, xu2023imagereward, kirstain2023pick, poesina2023iqpp, poesina_CVPR_2025}, which often address only specific tasks, \eg text-to-image generation \cite{kirstain2023pick,xu2023imagereward,yu2022scaling}, image captioning \cite{young-etal-2014-image}, query difficulty prediction \cite{poesina2023iqpp,poesina_CVPR_2025}, etc. For the evaluation of text-to-image generation, several datasets rely on human judgments. Pick-a-Pic~\cite{kirstain2023pick} collects human preferences over pairs of generated images from real user prompts, while ImageReward~\cite{xu2023imagereward} gathers expert rankings of generated images to train a reward model. PartiPrompts (P2)~\cite{yu2022scaling} introduces a curated set of English prompts, organized by category and difficulty, used for human evaluation of generated images. iQPP~\cite{poesina2023iqpp} is a benchmark that addresses query performance prediction for content-based image retrieval, whereas PQPP~\cite{poesina_CVPR_2025} introduces the first joint benchmark for predicting the difficulty of text prompts across two tasks: text-to-image generation and text-to-image retrieval.

English vision-language datasets have usually been employed as cornerstones for multilingual vision-language datasets, where retrieval and captioning have been key evaluation practices~\cite{thapliyal2022crossmodal,srinivasan2021wit,elliott2016multi30k}. Multi30K~\cite{elliott2016multi30k} extends Flickr30K~\cite{young-etal-2014-image} and provides captions for English and German, but only a subset of captions are parallel across languages. XTD10~\cite{aggarwal2021towards} includes 1{,}000 MS COCO captions in 11 languages for text-to-image retrieval. 
IGLUE~\cite{bugliarello-etal-2022-iglue} gathered several vision-language datasets across 20 languages, and WIT~\cite{srinivasan2021wit} is a large-scale Wikipedia-based text-image retrieval dataset covering 108 languages. XM3600~\cite{thapliyal2022crossmodal} contains 3{,}600 geographically diverse images with human-written captions across 36 languages. SeaCrowd \cite{2024-lovenia} is a data hub and benchmark suite for Southeast Asian (SEA) languages. For vision-language tasks, it offers localizations of the XM3600 \cite{thapliyal2022crossmodal} dataset in four SEA languages. 

In \cref{tab:related-datasets}, we compare \pamuk with existing datasets and benchmarks for vision-language tasks. In general, multilingual datasets usually cover one or (rarely) two of the five tasks included in \pamuk. Moreover, multilingual datasets that are large scale do not contain parallel prompts across languages, which hinders direct performance comparison between different languages, conflating performance drops caused by the language gap and those caused by the distinct data distribution. While XTD10~\cite{aggarwal2021towards} is a parallel multilingual resource, it covers a single task and only 1{,}000 parallel captions for 1{,}000 images, which is not sufficient for standard VLM fine-tuning. In summary, \pamuk is a unique resource, being the only one covering five vision-language tasks across seven languages (English included), enabling both intra- and inter-language performance assessments via parallel prompts and matching dataset splits across languages.

\noindent \textbf{Vision-language models.}
VLMs include models with joint (text and image) large-scale pre-training and models that combine vision encoders with LLMs. CLIP~\cite{radford2021learning} and ALIGN~\cite{jia2021scaling} are early models jointly trained via contrastive learning. Subsequent approaches improve either performance or efficiency using a broad range of enhancements, \eg sigmoid loss~\cite{zhai2023sigmoid}, image masking~\cite{Li_2023_CVPR}, data-efficient training~\cite{li2022supervision, mu2022slip}, faster inference~\cite{vasu2024mobile}, and carefully designed training techniques~\cite{sun2023eva}. CoCa~\cite{yu2022coca} and BLIP~\cite{li2022blip} combine contrastive learning with a generative objective, which further improves generalization capacity. General-purpose VLMs are often constructed by combining large-scale vision models with LLMs. BLIP-2~\cite{Li-ICML-2023} uses a Q-Former to bridge the gap between a frozen image encoder and a frozen LLM. LLaVA~\cite {liu2023visual} uses a linear projection layer to map visual features into the embedding space of an LLM, while Instruct-BLIP~\cite{dai2023instructblip} and LLaVA-Instruct~\cite{liu2023visual} use instruction tuning. Our contribution is to show that pre-trained models often fail on languages other than English, and that the performance gap can be recovered via language-specific fine-tuning.

\vspace{-0.1cm}
\section{Dataset}
\label{sec:data}
\vspace{-0.2cm}

\pamuk is a novel parallel multilingual dataset for VLU, which covers seven languages: Arabic, Danish, English, French, Hindi, Italian, Romanian. It builds upon an existing benchmark for prompt difficulty prediction, called PQPP \cite{poesina_CVPR_2025}, which targets the task of predicting the difficulty of prompts (a.k.a.~queries) in text-to-image generation and retrieval, where the prompts are only written in English. We extend PQPP in two fundamental ways. First, we expand the set of vision-language tasks to include text-to-image generation, image captioning, content-based image retrieval, along with the original prompt difficulty prediction for image generation/retrieval, therefore obtaining a more comprehensive set for evaluating vision-language understanding. Second, we collect parallel multilingual prompt translations in six additional languages via a semi-automatic process, covering a wide variety of languages from different families (Germanic, Indo-Aryan, Romance and Semitic). Thus, \pamuk enables the development and evaluation of vision-language models across different languages, as well as the direct comparison across languages, due its parallel nature.

\vspace{-0.1cm}
\subsection{Harnessing existing data and annotations}
\vspace{-0.1cm}

PQPP \cite{poesina_CVPR_2025} includes 10,200 prompts sourced from two existing datasets, namely Microsoft COCO (MS COCO) \cite{lin_LNSC_2014} and DrawBench \cite{saharia_NEURIPS_2022}. All 200 prompts from DrawBench are included in PQPP, while the other 10,000 are captions sampled from MS COCO via k-means clustering over sentence-BERT embeddings \cite{Thakur-NAACL-2020}. For each prompt, PQPP provides image retrieval relevance judgments for MS COCO images, as well as image generation relevance judgments for two images generated with Stable Diffusion XL (SDXL) \cite{Podell-ICLR-2024}, and another two generated with GLIDE \cite{Nichol-ICML-2021}. 

\vspace{-0.1cm}
\subsection{Collecting parallel multilingual annotations}
\label{sec:collecting}
\vspace{-0.1cm}

To diminish manual translation labor, we employ a semi-automatic translation process, involving a series of neural MT tools. These tools are used to propose reasonable initial translations that serve as starting points for human annotators, thus saving effort.

\noindent
\textbf{Human annotator recruitment.} 
We recruited human annotators to correct the automatically translated prompts. For each target language, the pool of reviewers contains between one and three people, each proficient in both English and their own target language. All annotators are adults who work in academia and hold at least a master or a PhD degree. The recruited annotators willingly agreed to engage in the annotation process, after consenting to the proposed terms. The main authors provided accurate and complete instructions regarding the annotation task. All active annotators received payment though their institution, around the average salaries in their home countries.


\noindent
\textbf{Annotation interface and instructions.}
We provided the annotators with (i) a document with annotation guidelines and examples, (ii) an annotation tool to streamline the review process, and (iii) a set of draft translations. In the annotation guidelines, the reviewers are asked to correct the translations if they are wrong, unclear or unnatural, and to preserve the tone and level of formality of the original source. We indicate that the text and not the associated picture is to be treated as authoritative. The guidelines instruct the reviewers to prefer fidelity to the original English prompt instead of language correctness or relevance to the image, replicating the same style/phrasing of the English version, while using the image for context only. We emphasize that \emph{fidelity to the original English prompt} is necessary to preserve the already available annotations imported from PQPP \cite{poesina_CVPR_2025}, \ie generation/retrieval relevance judgments, and prompt difficulty scores for generation/retrieval.
Hence, no elements present in the image and absent in the original English prompt should be added. The guidelines particularly mention that misspelled DrawBench prompts (which are introduced on purpose) must be translated using similar typos. To highlight the intentional misspellings from DrawBench, they are specifically tagged as \emph{$<$misspelled$>$} in the draft translation. If multiple source words collapse to one word in the target language (\eg ``sit'' and ``lye'' are both translated to ``st\u{a}'' in Romanian), a disambiguating expression can be used (\eg ``st\u{a} \^{i}n picioare'' vs.~``st\u{a} \^{i}ntins'').

\begin{figure}[t]
\centering
\includegraphics[width=0.9\columnwidth]{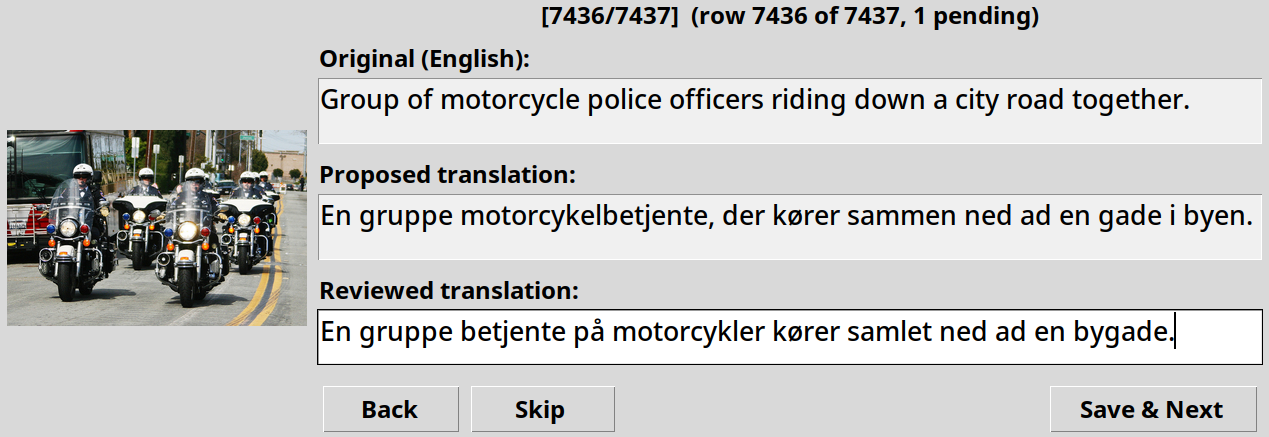}
\vspace{-0.33cm}
\caption{A screenshot of the translation tool used for manual reviewing. For each case, the tool displays the source prompt (in English), a draft prompt (in a target language) obtained via MT, and a text field where the reviewer can correct or entirely replace the draft prompt. When available, a relevant image from MS COCO is displayed to ease annotation.}
\label{fig:translation_tool}
\vspace{-0.22cm}
\end{figure}

As illustrated in \cref{fig:translation_tool}, the annotation tool displays one prompt at a time. For each case, it shows the original English prompt, a relevant image from MS COCO, the draft translation (obtained via MT), and an editable field where the annotator can correct or replace the draft translation. Each translation is automatically saved in the database. Reviewers can also choose to go back to correct a past translation, or skip the current translation and do it later. A reviewer can interrupt the annotation process at any time, and restart it from where they left.

\begin{table}[t]
\caption{Pairwise agreement among Romanian annotators, denoted as A1, A2, A3, on 100 control prompts. \emph{Identical} reports the percentage of identical prompts. For reference, the mean similarity between non-matching prompts is $0.28$ for Sentence-BERT and $0.47$ for COMET, respectively.}
\label{tab:reviewer-agreement}
\renewcommand{\arraystretch}{1.1}
\vspace{-0.35cm}
\setlength{\tabcolsep}{2.1pt}
\centering
\fontsize{7.3}{7.5}\selectfont{
\begin{tabular}{|l|ccc|c|}
\hline
Measure & A1--A2 & A1--A3 & A2--A3 & Average \\
\hline
Sentence-BERT & 0.987\std{0.029} & 0.990\std{0.025} & 0.984\std{0.030} & 0.987\std{0.028} \\
COMET & 0.926\std{0.066} & 0.934\std{0.064} & 0.936\std{0.064} & 0.932\std{0.065} \\
Identical & 49\% & 45\% & 32\% & 42\% \\
\hline
\end{tabular}
}
\vspace{-0.4cm}
\end{table}

\noindent
\textbf{Manual translation to Romanian.} Our largest pool of annotators are proficient Romanian speakers. Hence, for the Romanian language, all prompts are reviewed by at least one human annotator, starting from proposals produced using the Claude sonnet 4.6 API \cite{claude} with the following prompt \emph{``Translate the following English text to Romanian. Reply with ONLY the translation, nothing else.''}, followed by the original English text. 
The dataset is divided in three parts, and each part is reviewed by one of the three Romanian evaluators. A control subset of 100 randomly chosen prompts are included in all the three splits, in order to assess inter-reviewer agreement. We report the similarity of the manual translations provided by different reviewers for the control prompts in \cref{tab:reviewer-agreement}. The average similarity scores among annotators, measured via COMET \cite{rei2022comet} and Sentence-BERT \cite{reimers-2019-sentence-bert}, are consistently high, indicating a strong agreement among annotators.  




\noindent
\textbf{Harnessing MT models to reduce human effort.} Upon completing the manual annotation of Romanian prompts, we proceed to the next translation stage, which involves filtering translation proposals based on a voting scheme that employs three different MT models: DeepL \cite{deepl}, Google Translate \cite{googletranslate}, and Claude sonnet \cite{claude}. DeepL and Google Translate are chosen because they consistently reach top performance levels in MT evaluations reported by independent studies \cite{kamaluddin_jeef_2024, rivera_lre_2022}. Claude sonnet 4.6 is chosen because of its generic language modeling capabilities and its affordable pricing model. To calibrate our voting scheme, we first select a subset of 200 prompts, that are manually translated to French by one of the recruited annotators. Each MT model is asked to translate the subset of original English prompts and manually curated Romanian prompts to French. We further employ a voting scheme based on the number of translations that are in perfect agreement, where we vary the agreement threshold. A lower threshold reduces more of the manual annotation labor, at the cost of having low quality translations. In contrast, a higher threshold can put more translations through manual annotation, leading to high quality translations, but with a higher effort. We calibrate the threshold representing the minimum number of translations that must be in agreement via manual annotation on the selected subset of 200 prompts, where French is chosen as the target language. We select the smallest threshold for which the voted translation is in perfect agreement with the French annotator. The optimal threshold, minimizing annotation effort while ensuring perfect agreement, is three. Since there are six translations in total (three models $\times$ two source languages), there are cases where two different translations pass the minimum threshold. In such cases, we choose the one which includes the DeepL translation sourced in English (this rule is established during calibration on the 200 French prompts).

\begin{table}[t]
  \caption{Distribution of prompts that undergo manual review vs.~prompts that are saved from manual review via a voting scheme based on three translation models applied on English and Romanian source prompts. All Romanian prompts pass through rigorous manual review. The saved effort is reported in percentage points from the total number of prompts (10,200 per language).}
  \label{tab:review_savings}
  \vspace{-0.35cm}
  \renewcommand{\arraystretch}{1.1}
  \centering
  \fontsize{7.3}{7.5}\selectfont{
  \begin{tabular}{|l|r|r|r|}
    \hline
    Language & \#HT Prompts & \#MT Prompts & Saved Effort \\
    \hline
    Arabic  & 8{,}213 & 1{,}987 & 19.48\% \\
    Danish  & 7{,}434 & 2{,}766 & 27.12\% \\
    French  & 7{,}223 & 2{,}977 & 29.19\% \\
    Hindi   & 9{,}234 & 966 & 9.47\% \\
    Italian   & 7{,}201 & 2{,}999 & 29.40\% \\
    Romanian  & 10{,}200 & 0 & 0\% \\
    \hline
    Total & 49{,}505 & 11{,}695 & 19.11\% \\
    \hline
  \end{tabular}
  }
  \vspace{-0.15cm}
\end{table}

After calibrating the threshold for the proposed voting scheme, we apply it over the same set of MT models for the remaining languages (including French). The filtering leads to various human effort savings across languages, as reported in \cref{tab:review_savings}. Romance and Germanic languages exhibit higher labor reduction (nearly $30\%$), essentially because each of the source languages (English and Romanian) belongs to one of these language groups.



 

\begin{table}[t]
    \caption{Pairwise agreement between human and filtered machine translations on 100 control prompts. For reference, the mean similarity between non-matching prompts is $0.28$ for Sentence-BERT and $0.47$ for COMET, respectively.}
  \label{tab:bert_normal_vs_control}
  \vspace{-0.35cm}
  \renewcommand{\arraystretch}{1.1}
    \setlength{\tabcolsep}{3.6pt}
  \centering
  \fontsize{7.3}{7.5}\selectfont{
    \begin{tabular}{|l|ccccc|}
    \hline
    Measure & Arabic & Danish & French & Hindi & Italian \\
    \hline    
    Sentence-BERT  & 0.989 & 0.999 & 0.997 & 0.999 & 1.000 \\
    COMET  & 0.964 & 0.974 & 0.959 & 0.947 & 0.968 \\
    \hline
  \end{tabular}
  }
  \vspace{-0.4cm}
\end{table}

\noindent
\textbf{Manual translation to remaining languages.}
All prompts for which there is no agreement among at least three out of six automatic translations undergo manual review by proficient language speakers in each target language. Although we recruited proficient language speakers, we still inserted a subset of 100 control prompts per language from those filtered out by our voting scheme. The control set of prompts is used to verify the quality of the manual annotations, upon completing the annotation task. In \cref{tab:bert_normal_vs_control}, we report the similarities between control and manually labeled prompts, in terms of both Sentence-BERT and COMET measures. The values are roughly similar to the inter-annotator agreements reported in \cref{tab:reviewer-agreement}, confirming that the final translated prompts are of generally high quality. The total manual annotation time is estimated at 80 workdays (640 hours).



\vspace{-0.1cm}
\subsection{Dataset content and statistics}
\vspace{-0.1cm}

\pamuk contains 71,400 prompts that are grouped into 10,200 tuples of seven prompts. Each tuple contains the same prompt in seven different languages. Therefore, \pamuk is a fully parallel corpus, enabling exact comparative assessment across languages. Each prompt tuple is associated with two difficulty scores, one reflecting image generation and one reflecting image retrieval. In addition, it comes with four generated images: two from SDXL \cite{Podell-ICLR-2024}, and two from GLIDE \cite{Nichol-ICML-2021}, each annotated with relevance judgments. Retrieval relevance judgments for MS COCO images are also included for each prompt, totaling 530{,}360 relevance judgments. The associated images and annotations are inherited from PQPP \cite{poesina_CVPR_2025}. \pamuk enables their usage across multiple languages.
In \cref{tab:statistics}, we report token statistics for each language. English and Danish typically use fewer words than Italian and Romanian, while Arabic is shorter, mostly due to the fact that some articles are either presumed or being part of the target word. In contrast, Hindi generally uses more words.


\begin{table}[t]
  \caption{Token statistics across languages, including the maximum and median number of tokens per prompt, as well as the total number of tokens per language. DrawBench contains a few one word prompts, so the minimum tokens per prompt is always $1$.}
  \label{tab:statistics}
  \vspace{-0.35cm}
  \renewcommand{\arraystretch}{1.1}
      \setlength{\tabcolsep}{2.0pt}
  \centering
  \fontsize{7.3}{7.5}\selectfont{
    \begin{tabular}{|l|ccccccc|}
    \hline
    Length & English & Arabic & Danish & French & Hindi & Italian & Romanian \\
    \hline
    Max & 51 & 45 & 52 & 64 & 67 & 65 & 59 \\
    Median & 10 & 8 & 10 & 11  & 11  & 11 & 10 \\
    Total & 108{,}171 & 86{,}476 & 103{,}864 & 118{,}714 & 122{,}086 & 117{,}184 & 112{,}633 \\
    \hline
  \end{tabular}
  }
  \vspace{-0.4cm}
\end{table}

\vspace{-0.1cm}
\subsection{Tasks and performance measures}
\vspace{-0.1cm}

We propose a set of five tasks to evaluate vision-language understanding capabilities of foundation models. Across all tasks and languages, we employ a single official split of the prompts: 6,080 for training, 2,040 for validation, and 2,080 for testing. By using the same split across our collection of parallel prompts, we enable cross-lingual performance analysis. For each of the proposed tasks, we report multiple evaluation measures, since there is no agreement upon gold standard performance metrics.

\noindent 
\textbf{Text-to-image generation.} Given a text prompt in a certain language, the text-to-image generation task is to generate an image aligned with the input prompt. We evaluate the quality of the generated images via four evaluation metrics that measure alignment and human preference: HPSv2.1 \cite{wu_arxiv_2023}, HPSv3 \cite{ma_ICCV_2025}, PickScore \cite{kirstain2023pick} and CLIPScore \cite{hessel2021clipscore}.

\noindent 
\textbf{Image captioning.} Given an image as input, the image captioning task refers to generating a short textual description (caption) of the respective image. We score each caption against the reference caption (prompt) in the same language, using four evaluation metrics that consider character-level, word-level or semantic alignment:  spBLEU~\cite{goyal2022flores}, chrF++~\cite{popovic2017chrf}, BERTScore~\cite{zhang2020bertscore} and COMET~\cite{rei2022comet}.


\noindent 
\textbf{Content-based image retrieval.} Given a collection of images and a text query in a certain language, the content-based image retrieval task is to retrieve and rank the images according to their relevance to the input query. We evaluate systems using P@10, R@10, MRR (mean reciprocal rank) and nDCG@10, following common practice \cite{aggarwal2021towards, nakata2024rethinking}.

\noindent
\textbf{Prompt difficulty prediction (for generation).} Given a text prompt in a certain language, the prompt difficulty prediction task is to estimate how well a text-to-image generation model is expected to generate an image aligned with the input prompt.
We report the correlation of the predicted values with the Human Based Performance Prediction (HBPP) \cite{poesina_CVPR_2025} for GLIDE \cite{Nichol-ICML-2021} and SDXL \cite{Podell-ICLR-2024}. The correlation is measured using the Pearson $\rho$ and Kendall $\tau$ coefficients, following Poesina \etal~\cite{poesina_CVPR_2025}. 

\noindent
\textbf{Query difficulty prediction (for retrieval).} Given an image collection and a text query in a certain language, the query difficulty prediction task is to estimate, without accessing relevance judgments, how effectively an image retrieval system is expected to retrieve images relevant to the input query. We evaluate models based on how well they correlate with P@10 and reciprocal rank (RR), respectively. The target scores are computed for two retrieval systems, CLIP~\cite{radford2021learning} and BLIP-2~\cite{Li-ICML-2023}. The correlation is measured using both Pearson $\rho$ and Kendall $\tau$ coefficients. 

\vspace{-0.1cm}
\section{Experiments}
\label{sec:experiments}
\vspace{-0.2cm}

We conduct experiments on \pamuk to answer the following research questions: \emph{RQ1: What is the  VLU parity gap for English vs.~other languages?}, and \emph{RQ2: Can manually curated language-specific training data be used to close the gap?}. We evaluate baseline models in both zero-shot and fine-tuning setups to: (i) measure the performance gap between English and the six additional languages, and (ii) assess how this gap changes after fine-tuning on our data. 

\begin{table}[t]
  \caption{Text-to-image generation with SD 1.5 \cite{rombach2022high}, SD 3.5 \cite{esser2024scaling}, Scale-RAE \cite{tong2026scaling} and PixArt \cite{chen2024pixart}, on \pamuk. Models evaluated in three settings: ZS (En only) -- zero-shot on English prompts; ZS (lang.~avg.) -- zero-shot, averaged over all languages besides English; FT (lang.~avg.) -- fine-tuning, averaged over all languages besides English. Best language averages in bold.}
  \label{tab:t2i_overall}
  \vspace{-0.35cm}
  \centering
  \renewcommand{\arraystretch}{1.1}
  \setlength{\tabcolsep}{1.9pt}
  \fontsize{7.3}{7.5}\selectfont{
  \begin{tabular}{|l|l|c|c|c|c|}
    \hline
    Model & Setting & {HPSv2.1} & {HPSv3} & {PickScore} & {CLIPScore} \\
    \hline
    \multirow{3}{*}{{SD 1.5}} &
    ZS (En only) & 26.00 & 6.03 & 21.42 & 26.57 \\
    & ZS (lang.~avg.) & 16.98\std{3.97} & -2.66\std{3.37} & 18.41\std{1.58}  & 14.33\std{6.92} \\
    & FT (lang.~avg.) & \textbf{20.77}\std{2.98} & \textbf{0.25}\std{3.31} & \textbf{19.61}\std{1.02} & \textbf{20.52}\std{4.09} \\
    
    \hline
    \multirow{3}{*}{{SD 3.5}} &
    ZS (En only) & 28.12 & 10.37 & 22.39 & 27.06 \\
    & ZS (lang.~avg.) & 19.47\std{6.15} & 0.60\std{6.25} & 19.37\std{2.19} & 16.83\std{7.93} \\
    & FT (lang.~avg.) & \textbf{24.02}\std{2.88} & \textbf{5.68}\std{3.27} & \textbf{20.93}\std{1.00} & \textbf{22.39}\std{3.54} \\
    
    \hline
    \multirow{3}{*}{{Scale-RAE}} &
    ZS (En only) & 26.84 & 4.01 & 21.57 & 26.57 \\
    & ZS (lang.~avg.) & 19.50\std{3.75} & -2.58\std{3.23} & 19.06\std{1.28} & 17.06\std{4.79} \\
    & FT (lang.~avg.) & \textbf{26.23}\std{0.22} & \textbf{3.57}\std{0.21} & \textbf{21.31}\std{0.08} & \textbf{25.43}\std{0.34} \\
    
    \hline
    \multirow{3}{*}{{PixArt}} &
    ZS (En only) & 29.67 & 11.30 & 22.44 & 26.25 \\
    & ZS (lang.~avg.) & 21.13\std{6.12} & 2.57\std{6.22} & 19.38\std{2.38} & 15.60\std{8.04} \\
    & Fine-tune & \textbf{28.07}\std{1.73} & \textbf{9.93}\std{1.52} & \textbf{21.77}\std{0.66} & \textbf{23.77}\std{2.26} \\
    \hline
  \end{tabular}
  }
  \vspace{-0.15cm}
\end{table}

\begin{table}[t]
  \caption{Image captioning with BLIP-2 \cite{Li-ICML-2023}, Centurio \cite{geigle2025centurio}, LLaVA-1.5 \cite{liu2024improved}, Pangea-7B \cite{yue2025pangea}, on \pamuk. Models evaluated in three settings: ZS (En only) -- zero-shot on English prompts; ZS (lang.~avg.) -- zero-shot, averaged over all languages besides English; FT (lang.~avg.) -- fine-tuning, averaged over all languages besides English. Best language averages in bold.}
  \label{tab:captioning_overall}
  \vspace{-0.35cm}
  \centering
  \renewcommand{\arraystretch}{1.1}
  \setlength{\tabcolsep}{2.2pt}
  \fontsize{7.3}{7.5}\selectfont{
  \begin{tabular}{|l|l|c|c|c|c|}
    \hline
    Model & Setting & {spBLEU} & {chrF++} & {BERTScore} & {COMET} \\
    \hline
    \multirow{3}{*}{{BLIP-2}} &
    ZS (En only) & 12.7 & 31.9 & 0.771 & 0.681 \\
    & ZS (lang.~avg.) & 0.1\std{0.1} & 7.5\std{5.8} & 0.684\std{0.028} & 0.428\std{0.077} \\
    & FT (lang.~avg.) & \textbf{15.0}\std{1.0} & \textbf{29.5}\std{3.3} & \textbf{0.786}\std{0.007} & \textbf{0.657}\std{0.048} \\
    \hline
    \multirow{3}{*}{{Centurio}} &
    ZS (En only) & 11.5 & 30.8 & 0.799 & 0.685 \\
    & ZS (lang.~avg.) & 9.4\std{1.5} & 24.6\std{4.2} & 0.775\std{0.013} & 0.607\std{0.054} \\
    & FT (lang.~avg.) & \textbf{14.7}\std{1.3} & \textbf{29.9}\std{3.3} & \textbf{0.790}\std{0.006} & \textbf{0.663}\std{0.045} \\
    \hline
    \multirow{3}{*}{{LLaVA}} &
    ZS (En only) & 2.1 & 21.4 & 0.694 & 0.629 \\
    & ZS (lang.~avg.) & 1.2\std{1.0} & 15.5\std{6.4} & 0.666\std{0.035} & 0.464\std{0.130} \\
    & FT (lang.~avg.) & \textbf{14.2}\std{0.5} & \textbf{28.7}\std{2.9} & \textbf{0.789}\std{0.007} & \textbf{0.655}\std{0.038} \\
    \hline
    \multirow{3}{*}{{Pangea}} &
    ZS (En only) & 0.8 & 12.4 & 0.647 & 0.594 \\
    & ZS (lang.~avg.) & 1.7\std{0.7} & 15.6\std{2.6} & 0.662\std{0.014} & 0.617\std{0.042} \\
    & FT (lang.~avg.) & \textbf{15.9}\std{0.8} & \textbf{30.6}\std{2.8} & \textbf{0.795}\std{0.006} & \textbf{0.669}\std{0.034} \\
    \hline
  \end{tabular}
  }
  \vspace{-0.4cm}
\end{table}

\begin{figure*}[t]
  \centering
  \includegraphics[width=0.83\textwidth]{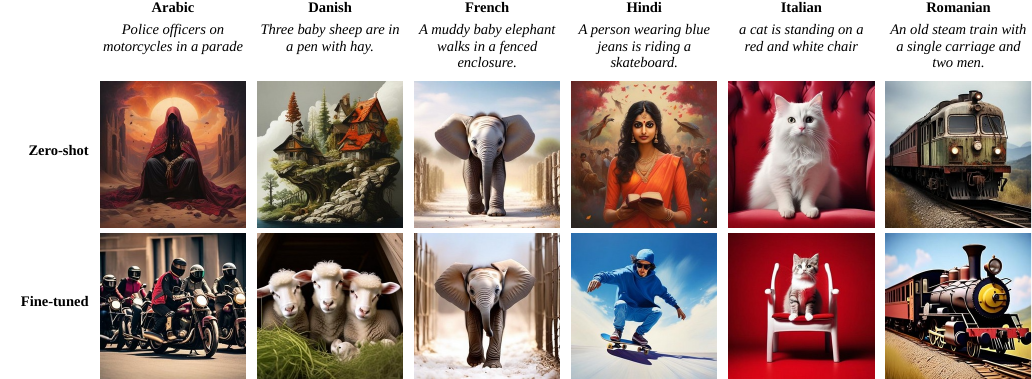}
  \vspace{-0.32cm}
  \caption{Text-to-image generation examples (one target language per prompt) for PixArt-Sigma before and after fine-tuning on \pamuk. Only the original English prompts are shown, but the model is prompted with the corresponding translated versions.}
  \label{fig:t2i_examples}
  \vspace{-0.2cm}
\end{figure*}

\begin{figure*}[t]
  \centering
  \includegraphics[width=0.83\textwidth]{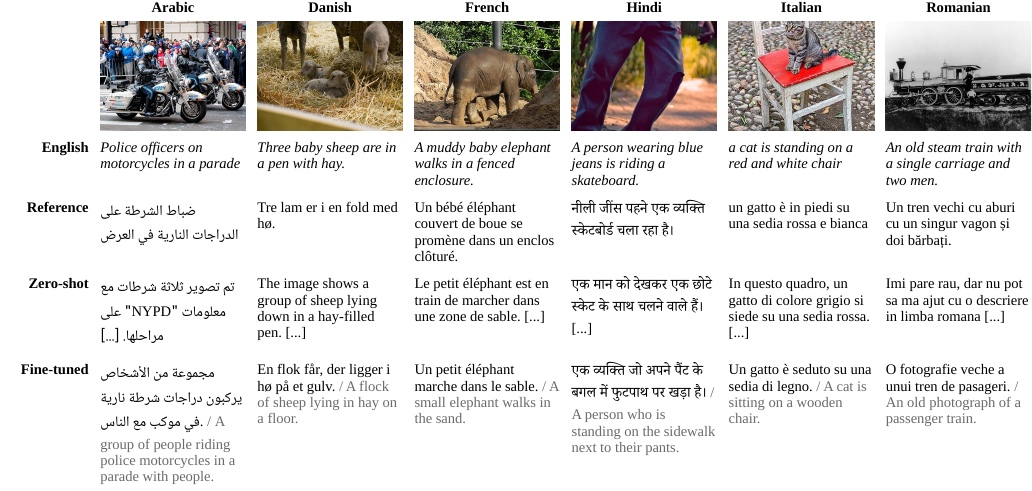}
  \vspace{-0.35cm}
  \caption{Image captioning examples (one target language per image) for LLaVA before and after fine-tuning on \pamuk. Long zero-shot captions are truncated ([...]), the complete outputs being shown in \cref{fig:captioning_examples_full}. English translations of captions returned by the fine-tuned LLaVA follow the slash (in gray).}
  \label{fig:captioning_examples}
  \vspace{-0.4cm}
\end{figure*}

\vspace{-0.1cm}
\subsection{Baseline models}
\vspace{-0.1cm}


\noindent 
\textbf{Text-to-image generation.}
For text-to-image generation, we consider four state-of-the-art generation models: Stable Diffusion 1.5 (SD 1.5)~\cite{rombach2022high}, Stable Diffusion 3.5 Medium (SD 3.5)~\cite{esser2024scaling}, PixArt-Sigma~\cite{chen2024pixart}, and Scale-RAE~\cite{tong2026scaling}.

\noindent 
\textbf{Image captioning.}
For image captioning, we consider four models that differ in their architecture and exposure to multilingual data: BLIP-2 6.7B~\cite{Li-ICML-2023}, Centurio~\cite{geigle2025centurio}, LLaVA-1.5 7B~\cite{liu2024improved}, and Pangea-7B~\cite{yue2025pangea}.

\noindent 
\textbf{Content-based image retrieval.}
For content-based image retrieval, we experiment with CLIP \cite{li2022supervision} and BLIP-2 6.7B~\cite{Li-ICML-2023}. Since the original CLIP uses an English text encoder, we replace it with a multilingual XLM-R encoder \cite{conneau-etal-2020-unsupervised}, and report results with both versions of CLIP.

\noindent
\textbf{Prompt/query difficulty prediction.}
We use the same selection of models for both prompt and query difficulty prediction. Following Poesina \etal \cite{poesina_CVPR_2025}, we use one pre-generation/retrieval predictor based on BERT \cite{Devlin-NAACL-2019}, and two post-generation/retrieval predictors, Long-CLIP \cite{Zhang-ECCV-2024} and Correlation CNN \cite{sun2018assessing}. Whenever possible, we replace English text encoders with multilingual counterparts.

\begin{table}[t]
\caption{Content-based image retrieval with CLIP \cite{li2022supervision}, XLM-R-based CLIP \cite{conneau-etal-2020-unsupervised}, and BLIP-2 \cite{Li-ICML-2023} on \pamuk. Models evaluated in three settings: ZS (En only) -- zero-shot on English prompts; ZS (lang.~avg.) -- zero-shot, averaged over all languages besides English; FT (lang.~avg.) -- fine-tuning, averaged over all languages besides English. Best language averages in bold.}
\vspace{-0.35cm}
\renewcommand{\arraystretch}{1.1}
\setlength{\tabcolsep}{2.0pt}
\centering
\fontsize{7.3}{7.5}\selectfont{
\begin{tabular}{|l|l|c|c|c|c|}
\hline
Model & Setting & P@10 & R@10 & MRR & nDCG@10 \\

 \hline
 \multirow{3}{*}{{CLIP}} 
 & ZS (En only) & 0.177 & 0.328 & 0.471 & 0.356 \\
  & ZS (lang.~avg.) & 0.021\std{0.025} & 0.030\std{0.035} & 0.060\std{0.063} & 0.035\std{0.040} \\
  & FT (lang.~avg.) & \textbf{0.158}\std{0.038} & \textbf{0.218}\std{0.082} & \textbf{0.365}\std{0.096} & \textbf{0.266}\std{0.085} \\
 
 \hline
 \multirow{3}{*}{\shortstack[l]{XLM-R\\CLIP}} 
 & ZS (En only) & 0.206 & 0.361 & 0.552 & 0.420 \\
 & ZS (lang.~avg.) & 0.151\std{0.022} & 0.260\std{0.045} & 0.400\std{0.062} & 0.293\std{0.051} \\
 & FT (lang.~avg.) & \textbf{0.217}\std{0.009} & \textbf{0.346}\std{0.024} & \textbf{0.535}\std{0.033} & \textbf{0.409}\std{0.028} \\
 
 \hline
 \multirow{3}{*}{{BLIP-2}} 
 & ZS (En only) & 0.253 & 0.395 & 0.619 & 0.484 \\
 & ZS (lang.~avg.) & 0.023\std{0.026} & 0.036\std{0.041} & 0.064\std{0.066} & 0.039\std{0.043} \\
 &FT (lang.~avg.) & \textbf{0.144}\std{0.068} & \textbf{0.163}\std{0.104} & \textbf{0.304}\std{0.148} & \textbf{0.220}\std{0.121} \\

\hline
\end{tabular}
}
\label{tab:retrieval}
\vspace{-0.4cm}
\end{table}

\noindent
\textbf{Fine-tuning procedure.} For efficient adaptation, we use LoRA-based fine-tuning~\cite{hu2022lora}, which keeps the pre-trained weights frozen and trains only low-rank adapters. We tune all hyperparameters on the validation split, considering LoRA ranks between $8$ and $64$, learning rates between $10^{-5}$ and $10^{-3}$, weight decays between $0$ and $0.1$, and batch sizes between $2$ and $64$. Parameters are generally tuned per language. Training is based on AdamW \cite{loshchilov-iclr-2019-adamw}. More details about models and hyperparameters are given in \cref{supp_sec_models}.







\begin{table*}[t]
    \centering
    \caption{Prompt and query difficulty prediction with BERT \cite{Devlin-NAACL-2019}, Long-CLIP \cite{Zhang-ECCV-2024}, and Correlation CNN \cite{sun2018assessing} on \pamuk. Models evaluated in four settings: ZS (En only) -- zero-shot on English prompts, as reported in \cite{poesina_CVPR_2025}; ZS (En only, rep.) -- zero-shot on English prompts, reproduced with multilingual text encoders; ZS (lang.~avg.) -- zero-shot, averaged over all languages besides English; FT (lang.~avg.) -- fine-tuning, averaged over all languages besides English. Best language averages in bold.}
    \label{tab:difficulty_overall}
    \setlength{\tabcolsep}{1.95pt}
    \vspace{-0.35cm}
    \renewcommand{\arraystretch}{1.1}
    \fontsize{7.3}{7.5}\selectfont{
    \begin{tabular}{|l|l|c|c|c|c|| c|c|c|c| c|c|c|c|}
        \hline
        & & \multicolumn{4}{c||}{Generative Task}
          & \multicolumn{8}{c|}{Retrieval Task} \\
        \cline{3-6}
        \cline{7-14}

     Predictor & Setting & \multicolumn{2}{c}{HBPP\,/\,GLIDE}
          & \multicolumn{2}{c||}{HBPP\,/\,SDXL}
          & \multicolumn{2}{c|}{P@10\,/\,CLIP}
          & \multicolumn{2}{c|}{RR\,/\,CLIP}
          & \multicolumn{2}{c|}{P@10\,/\,BLIP-2}
          & \multicolumn{2}{c|}{RR\,/\,BLIP-2} \\

        \cline{3-4}
        \cline{5-6}
        \cline{7-8}
        \cline{9-10}
        \cline{11-12}
        \cline{13-14}

         & 
          & Pearson & Kendall
          & Pearson & Kendall
          & Pearson & Kendall
          & Pearson & Kendall
          & Pearson & Kendall
          & Pearson & Kendall \\
        \hline

        \multirow{4}{*}{\shortstack[l]{BERT\\(pre-gen./ret.)}}
          & ZS (En only)
          & .566 & .406 & .281 & .232 & .451 & .277 & .221 & .176 & .511 & .328 & .168 & .139 \\

          & ZS (En only, rep.)
          & .601 & .432 & .329 & .230 & .387 & .256 & .230 & .163 & .450 & .291 & .194 & .146 \\
        \cline{2-14}

          & ZS (lang.~avg.)
          & .502\std{.042} & .359\std{.031} & .197\std{.038} & .143\std{.016} & .228\std{.028} & .155\std{.022} & .170\std{.034} & .116\std{.024} & .261\std{.037} & .181\std{.027} & .083\std{.033} & .071\std{.024} \\

          & FT (lang.~avg.)
          & \textbf{.573}\std{.008} & \textbf{.411}\std{.007}
          & \textbf{.304}\std{.016} & \textbf{.217}\std{.017}
          & \textbf{.357}\std{.014} & \textbf{.228}\std{.009}
          & \textbf{.213}\std{.016} & \textbf{.156}\std{.014}
          & \textbf{.420}\std{.025} & \textbf{.280}\std{.017}
          & \textbf{.189}\std{.029} & \textbf{.139}\std{.021}  \\
        \hline

        \multirow{4}{*}{\shortstack[l]{Long-CLIP\\(post-gen./ret.)}}
          & ZS (En only) & .649 & .474 & .380 & .246 & .473 & .299 & .200 & .149 & .498 & .358 & .166 & .150 \\
          
          & ZS (En only, rep.) & .671 & .491 & .337 & .224 & .599 & .375 & .335 & .235 & .644 & .444 & .281 & .208 \\
        \cline{2-14}

          & ZS (lang.~avg.)
          & .666\std{.002} & \textbf{.487}\std{.002} & .319\std{.008} & .211\std{.007}
          & .526\std{.035} & .355\std{.018} & .424\std{.021} & .335\std{.029}
          & .547\std{.032} & .380\std{.028} & .134\std{.056}  &.089\std{.043}  \\

          & FT (lang.~avg.)
          & \textbf{.667}\std{.001} & \textbf{.487}\std{.001}
          & \textbf{.325}\std{.007} & \textbf{.212}\std{.007}
          & \textbf{.601}\std{.021} & \textbf{.391}\std{.024}
          & \textbf{.451}\std{.053} & \textbf{.356}\std{.059}
          & \textbf{.625}\std{.030} & \textbf{.412}\std{.024} &\textbf{.147}\std{.039} & \textbf{.101}\std{.027} \\
        \hline

        \multirow{4}{*}{\shortstack[l]{Corr.~CNN\\(post-gen./ret.)}}
          & ZS (En only) & .548 & .393 & .159 & .107 & .270 & .186 & .189 & .162 & .159 & .133 & .206 & .158 \\
 
 & ZS (En only, rep.) & .586 & .426 & .124 & .046 & .338 & .213 & .256 & .190 & .305 & .220 & .176 & .130 \\
        \cline{2-14}

          & ZS (lang.~avg.)
& .564\std{.011} & .404\std{.011}
& \textbf{.166}\std{.025} & \textbf{.108}\std{.030}
& \textbf{.263}\std{.092} & .145\std{.038}
& .174\std{.063} & .132\std{.041}
& .259\std{.066} & .176\std{.064}
& .122\std{.035} & .089\std{.023} \\

& FT (lang.~avg.)
& \textbf{.567}\std{.016} & \textbf{.409}\std{.014}
& .163\std{.031} & .092\std{.037}
& .255\std{.052} & \textbf{.157}\std{.031}
& \textbf{.197}\std{.042} & \textbf{.151}\std{.024}
& \textbf{.283}\std{.068} & \textbf{.198}\std{.072}
& \textbf{.157}\std{.069} & \textbf{.119}\std{.061} \\
        \hline
    \end{tabular}%
    }
    \vspace{-0.4cm}
\end{table*}

\vspace{-0.1cm}
\subsection{Main results}
\label{sec:exp_captioning}
\vspace{-0.1cm}

\noindent
\textbf{Text-to-image generation.}
We report text-to-image generation results aggregated across languages in \cref{tab:t2i_overall} (per-language results are in \cref{tab:t2i_full}). Across all metrics, performance degrades significantly for all models when English prompts are replaced with translated versions. Fine-tuning on our training data improves every model on every metric. Moreover, the gap is almost completely recovered for Scale-RAE and PixArt-Sigma. While fine-tuning is beneficial across all six languages, the gains for SD 3.5, PixArt-Sigma and Scale-RAE are generally higher for Arabic and Hindi, where the zero-shot models perform the worst (see \cref{tab:t2i_full}). In \cref{fig:t2i_examples}, we illustrate images generated by PixArt-Sigma, before and after fine-tuning. For several prompts (first, second and fourth columns), the zero-shot model is completely off, indicating significant disparities between zero-shot and fine-tuned counterparts.




\noindent
\textbf{Image captioning.}
We report the image captioning results aggregated across languages in \cref{tab:captioning_overall} (per-language results are in \cref{tab:captioning_full}). Most VLMs experience performance drops when captions need to be produced in a language other than English. Due to its multi-lingual pre-training, Pangea-7B is the only model obtaining better zero-shot performance on other languages than on English. Fine-tuning on our training data improves all four models across all four metrics. This is true even for the multilingual VLMs, namely Centurio and Pangea-7B.
The gains hold across all six languages, as shown by the per-language results in \cref{tab:captioning_full}. In \cref{fig:captioning_examples}, we showcase captions returned by LLaVA, before and after fine-tuning. The zero-shot LLaVA returns long and inaccurate descriptions, or even fails to answer in the target language, while the fine-tuned LLaVA returns precise and relevant captions. 

\noindent
\textbf{Content-based image retrieval.} In \cref{tab:retrieval}, we report aggregated retrieval results over all languages (per-language results are in \cref{tab:retrieval_metrics_full}). When changing the queries from English to another language, retrieval results consistently drop for all zero-shot models. Fine-tuning brings significant gains for all models. However, the performance gap w.r.t.~English is only restored (with marginal difference) from the XLM-R-based CLIP.

\noindent
\textbf{Generation difficulty prediction.}
We report the results for generation difficulty prediction, averaged over languages, in \cref{tab:difficulty_overall} (per-language results are in \cref{tab:difficulty_full}). The BERT predictor exhibits visible drops when the prompt is not in English, but the degradation is generally recovered via fine-tuning. Long-CLIP, which integrates the XLM-R encoder with multilingual capabilities, maintains its predictive performance in the zero-shot (language average) setting, experiencing only marginal degradations that are easily recovered via fine-tuning. Unlike the other models, the correlation CNN is not significantly affected by the language differences. This is due to its post-generation nature, focusing mostly on the generated images rather than the prompt. 

\noindent
\textbf{Retrieval difficulty prediction.}
Retrieval difficulty prediction results sit next to corresponding  generation difficulty prediction results in \cref{tab:difficulty_overall} (aggregated) and \cref{tab:difficulty_full} (per language), respectively. The zero-shot performance levels of the BERT and correlation CNN predictors degrade for non-English prompts, while the zero-shot Long-CLIP is more resilient to query language changes, as it integrates the multilingual XLM-R text encoder. Fine-tuning consistently improves all three query difficulty predictors.

\noindent
\textbf{Answering research questions.}
The results across the five vision-language tasks indicate that performance generally degrades by significant margins when prompts are not in English, even in cases where models integrate multilingual text encoders. Regarding \emph{RQ1}, we find that the language parity gap is usually large, across the majority of VLU tasks and VLMs. Regarding \emph{RQ2}, we observe that fine-tuning on \pamuk always reduces the performance parity gap between English and other languages, though the gap is not entirely closed in all cases.

\vspace{-0.1cm}
\section{Conclusion}
\vspace{-0.2cm}

In this paper, we proposed a novel parallel multilingual multitask benchmark for VLU tasks, called \pamuk. It comprises 10,200 prompts with parallel versions in seven languages, which are associated with natural and generated images, enabling a range of five vision-language tasks. The parallel nature of the prompts coupled with the same dataset split across languages, enables immediate intra-language and inter-language comparisons of various VLMs. In particular, we carried out experiments that confirm the presence of a large inter-language performance parity gap between English and other languages, across various zero-shot VLMs. We also showed that this gap can be reduced by significant margins via fine-tuning, though not always entirely.

In future work, we plan to expand the number of covered languages and the set of evaluated models.

\noindent
\textbf{Acknowledgment.} This research is supported by the project ``Romanian Hub for Artificial Intelligence - HRIA'', Smart Growth, Digitization and Financial Instruments Program, 2021-2027, MySMIS no.~351416. This research is also supported by two grants of the Ministry of Research, Innovation and Digitization, CNCS - UEFISCDI, project numbers PN-IV-P1-PCE-2023-0354 and PN-IV-P6-6.3-SOL-2024-2-0197, within PNCDI IV. The authors acknowledge the High Performance Computing of the InGeo Department for Italian language results.





{
    \small
    \bibliographystyle{ieeenat_fullname}
    \bibliography{main}
}

\clearpage
\maketitlesupplementary
\setcounter{page}{1}
\setcounter{figure}{0}
\setcounter{table}{0}
\setcounter{equation}{0}
\renewcommand{\thefigure}{S\arabic{figure}}
\renewcommand{\thetable}{S\arabic{table}}
\renewcommand{\theequation}{S\arabic{equation}}

\vspace{-0.1cm}
\section{Data Ethics and Licensing}
\vspace{-0.1cm}
\subsection{Ethical considerations}
\vspace{-0.1cm}

Computing power and therefore energy was used in the process of translating the texts and performing the experiments. This contributes to climate change. The effect was, however, small since the translations and experiments were relatively inexpensive in terms of computation, at least compared with the typical modern workload. Moreover the translation effort is likely to be reused since it remains embedded in the public dataset. 

The original English captions permit gender neutral forms for various words (like doctor), while Arabic, Hindi, Romanian and French often force gender choices, especially on plurals and pronouns. In this respect the initial machine translations are known to have implicit biases \cite{currey-EMNLP-2022}, which are propagated to our dataset. The Danish language is closer to English in this respect.

\vspace{-0.1cm}
\subsection{Licensing}
\vspace{-0.1cm}

We release the parallel multilingual prompts under the CC-BY 4.0 license\footnote{\scriptsize{\url{https://creativecommons.org/licenses/by/4.0/deed}}}. Existing data samples are not part of our distribution, and keep their original licensing terms.


\begin{table}[t]
  \caption{Training hyperparameters for all five tasks. LoRA configurations used for text-to-image generation and image captioning. The LoRA rank is denoted as $r$, the LoRA scaling factor as $\alpha$, and the peak learning rate as $\eta$. WD denotes weight decay.}
  \vspace{-0.3cm}
  \renewcommand{\arraystretch}{1.1}
  \label{tab:hyperparameters}
  \centering
  \setlength{\tabcolsep}{3pt}
  \fontsize{7.3}{7.5}\selectfont{
  \begin{tabular}{|l|l|c|c|c|c|c|}
    \hline
    Task & Model & $r$ & $\alpha$ & $\eta$ & Batch & Epochs \\
    \hline
    \multirow{4}{*}{\shortstack[l]{Text-to-image\\generation}} 
     & PixArt-Sigma & 16 & 16 & $5 {\cdot} 10^{-4}$ & 16 & 10 \\
     & SD 1.5 & 16 & 16 & $10^{-4}$ & 16 & 10 \\
     & SD 3.5 & 16 & 16 & $10^{-4}$ & 16 & 10 \\
     & Scale-RAE & 16 & 16 & $5 {\cdot} 10^{-4}$ & 16 & 10 \\
    \hline
    \multirow{4}{*}{\shortstack[l]{Image\\captioning}} & BLIP-2 & 32 & 64 & $2 {\cdot} 10^{-4}$ & 8 & 5 \\
     & Centurio & 8 & 16 & $10^{-4}$ & 2 & 2 \\
     & LLaVA-1.5 & 16 & 32 & $2 {\cdot} 10^{-4}$ & 2 & 2 \\
     & Pangea-7B & 16 & 32 & $2 {\cdot} 10^{-4}$ & 2 & 2 \\
         \hline
    \hline
    Task & Model & Optimizer & WD & $\eta$ & Batch & Epochs \\
        \hline  
    \multirow{3}{*}{\shortstack[l]{Image\\ retrieval}}
     & CLIP & AdamW & 0.1 & $5 {\cdot}10^{-6}$ & 48 & 15 \\
     & XLM-R CLIP & AdamW & 0.1 & $5 {\cdot}10^{-6}$ & 48 & 15 \\
     & BLIP-2 & AdamW & 0.01 & $5{\cdot}10^{-5}$ & 16 & 15 \\\hline     
    \multirow{3}{*}{\shortstack[l]{Generation\\difficulty}}
     & mBERT & AdamW & 0.01 & $5 {\cdot}10^{-5}$ & 256 & 15 \\
     & Long-CLIP & Adam & 0 & $5 {\cdot}10^{-5}$ & 256 & 100 \\
     & Corr.~CNN & Adam & 0 & $10^{-4}$ & 128 & 25 \\
    \hline
    \multirow{3}{*}{\shortstack[l]{Retrieval\\difficulty}}
     & mBERT & AdamW & 0.1 & $5 {\cdot}10^{-5}$ & 256 & 15 \\
     & Long-CLIP & AdamW & 0.01 & $5 {\cdot}10^{-5}$ & 256 & 25 \\
     & Corr.~CNN & Adam & 0 & $10^{-4}$ & 32 & 25 \\

    \hline
  \end{tabular}
  }
  \vspace{-0.3cm}
\end{table}

\begin{table*}[t]
\caption{Per-language text-to-image generation results with Scale-RAE, PixArt-Sigma, Stable Diffusion 1.5 and Stable Diffusion 3.5. HPSv2.1 ($\times$100), HPSv3, PickScore and CLIPScore, are all computed against the English caption, one image per prompt. PickScore is the scaled cosine similarity between the image and the caption, and CLIPScore is computed with CLIP ViT-L/14 and a weight of 100. The results are reported on the test split. For PixArt-Sigma and Stable Diffusion 3.5, the Arabic and Hindi prompts are romanized with the \texttt{uroman} tool. English and zero-shot are single runs. Scores of fine-tuned models represent the mean and standard deviation over three training seeds, with the same sampling seed for each prompt in every run. The average is over the six languages (English is excluded), and its deviation is taken over the three per-seed averages. The best score for each language among both zero-shot and fine-tune models is in bold.}
  \renewcommand{\arraystretch}{1.1}
  \label{tab:t2i_full}
  \vspace{-0.3cm}
  \centering
  \setlength{\tabcolsep}{4pt}
  \fontsize{7.3}{7.5}\selectfont{
  \begin{tabular}{|l|l|c|c|c|c|c|c|c|c|}
    \hline
    \multirow{2}{*}{Language} & \multirow{2}{*}{Model} & \multicolumn{2}{c|}{HPSv2.1} & \multicolumn{2}{c|}{HPSv3} & \multicolumn{2}{c|}{PickScore} & \multicolumn{2}{c|}{CLIPScore} \\
    \cline{3-10}
     &  & Zero-shot & Fine-tuned & Zero-shot & Fine-tuned & Zero-shot & Fine-tuned & Zero-shot & Fine-tuned \\
    \hline
    \hline
    \multirow{4}{*}{English} & Stable Diffusion 1.5 & 26.00 & -- & 6.03 & -- & 21.42 & -- & \textbf{27.07} & -- \\
     & Stable Diffusion 3.5 & 28.12 & -- & 10.37 & -- & 22.39 & -- & 27.06 & -- \\
     & Scale-RAE & 26.84 & -- & 4.01 & -- & 21.57 & -- & 26.57 & -- \\
     & PixArt-Sigma & \textbf{29.67} & -- & \textbf{11.30} & -- & \textbf{22.44} & -- & 26.25 & -- \\
     
    \hline
    \hline
    \multirow{4}{*}{Arabic} & Stable Diffusion 1.5 & 11.84 & 17.74\std{0.154} & $-$7.05 & $-$3.16\std{0.159} & 16.66 & 18.54\std{0.041} & 6.39 &  15.99\std{0.186} \\
     & Stable Diffusion 3.5 & 11.10 & 20.03\std{0.911} & $-$7.59 & 1.15\std{1.087} & 16.83 & 19.55\std{0.339} & 6.75 & 17.41\std{1.293} \\
     & Scale-RAE & 15.81 & \textbf{26.07}\std{0.013} & $-$5.72 & 3.45\std{0.037} & 17.85 & \textbf{21.25}\std{0.001} & 12.25 & \textbf{25.17}\std{0.062} \\
     & PixArt-Sigma & 12.78 & 24.70\std{0.818} & $-$6.54 & \textbf{6.97}\std{0.766} & 16.42 & 20.50\std{0.308} & 4.91 & 19.42\std{1.047} \\
     
     \hline
     \hline
    \multirow{4}{*}{Danish}  & Stable Diffusion 1.5 & 17.47 & 21.41\std{0.160} & $-$2.04 & 1.21\std{0.260} & 18.05 & 19.86\std{0.023} & 15.91 & 21.69\std{0.041} \\
     & Stable Diffusion 3.5 & 20.97 & 24.92\std{0.217} & 2.04 & 6.67\std{0.217} & 20.03 & 21.24\std{0.055} & 19.37 & 23.74\std{0.179} \\
     & Scale-RAE & 18.44 & 26.21\std{0.013} & $-$3.51 & 3.55\std{0.035} & 18.76 & 21.31\std{0.011} & 16.46 & \textbf{25.41}\std{0.038} \\
     & PixArt-Sigma & 23.46 & \textbf{28.86}\std{0.035} & 5.20 & \textbf{10.61}\std{0.037} & 20.47 & \textbf{22.06}\std{0.017} & 19.19 & 24.76\std{0.074} \\
    
     \hline
     \hline
    \multirow{4}{*}{French} & Stable Diffusion 1.5 & 21.83 & 24.07\std{0.066} & 1.63 & 3.85\std{0.151} & 20.23 & 20.75\std{0.032} & 22.25 & 24.95\std{0.018} \\
     & Stable Diffusion 3.5 & 26.25 & 27.01\std{0.148} & 7.86 & 9.12\std{0.205} & 21.69 & 21.97\std{0.022} & 25.08 & 25.84\std{0.101} \\
     & Scale-RAE & 24.67 & 26.48\std{0.039} & 2.02 & 3.79\std{0.042} & 20.80 & 21.42\std{0.008} & 23.47 & \textbf{25.88}\std{0.032} \\
     & PixArt-Sigma & 27.38 & \textbf{29.23}\std{0.020} & 8.87 & \textbf{10.96}\std{0.044} & 21.67 & \textbf{22.24}\std{0.013} & 23.47 & 25.37\std{0.057} \\
     
     \hline
     \hline
    \multirow{4}{*}{Hindi} & Stable Diffusion 1.5 & 13.04 & 16.69\std{0.085} & $-$5.83 & $-$4.34\std{0.031} & 16.42 & 18.26\std{0.013} & 5.75 & 15.16\std{0.066} \\
     & Stable Diffusion 3.5 & 13.05 & 20.99\std{0.081} & $-$6.03 & 2.26\std{0.177} & 16.55 & 19.87\std{0.052} & 7.35 & 18.58\std{0.081} \\
     & Scale-RAE & 15.33 & 25.92\std{0.038} & $-$6.10 & 3.30\std{0.058} & 17.56 & 21.21\std{0.006} & 11.36 & \textbf{24.98}\std{0.057} \\
     & PixArt-Sigma & 14.42 & \textbf{27.80}\std{0.100} & $-$3.50 & \textbf{9.73}\std{0.118} & 16.37 & \textbf{21.62}\std{0.029} & 6.22 & 23.20\std{0.080} \\
     
     \hline
     \hline
    \multirow{4}{*}{Italian} & Stable Diffusion 1.5 & 20.57 & 23.39\std{0.093} & 0.25 & 3.08\std{0.058} & 19.81 & 20.51\std{0.021} & 20.34 &  24.08\std{0.155} \\
     & Stable Diffusion 3.5 & 24.75 & 26.49\std{0.251} & 5.95 & 8.52\std{0.289} & 21.20 & 21.77\std{0.078} & 23.48 & 25.34\std{0.208} \\
     
     & Scale-RAE & 22.88 & 26.48\std{0.005} & 0.24 & 3.84\std{0.015} & 20.20 & 21.40\std{0.003} & 21.22 & \textbf{25.74}\std{0.042} \\
     & PixArt-Sigma & 26.21 & \textbf{29.14}\std{0.048} & 7.65 & \textbf{10.91}\std{0.044} & 21.29 & \textbf{22.19}\std{0.013} & 22.05 & 25.22\std{0.032} \\
     
     \hline
     \hline
    \multirow{4}{*}{Romanian} & Stable Diffusion 1.5 & 17.15 & 21.35\std{0.069} & $-$2.93 & 0.83\std{0.103} & 18.68 &19.79\std{0.007} & 15.39 &  21.30\std{0.017} \\
     & Stable Diffusion 3.5 & 20.72 & 24.67\std{0.994} & 1.40 & 6.36\std{1.372} & 19.93 & 21.17\std{0.338} & 18.97 & 23.42\std{0.938} \\
     & Scale-RAE & 19.87 & 26.19\std{0.023} & $-$2.38 & 3.51\std{0.035} & 19.16 & 21.31\std{0.005} & 17.61 & \textbf{25.40}\std{0.023} \\
     & PixArt-Sigma & 22.56 & \textbf{28.66}\std{0.046} & 3.76 & \textbf{10.39}\std{0.054} & 20.03 & \textbf{22.02}\std{0.019} & 17.74 & 24.64\std{0.017} \\
     
    \hline
    \hline
    \multirow{4}{*}{Average} & Stable Diffusion 1.5 & 16.98 & 20.77\std{0.092} & $-$2.66 & 0.25\std{0.099} & 18.41 & 19.61\std{0.023} & 14.33 & 20.52\std{0.080} \\
     & Stable Diffusion 3.5 & 19.47 & 24.02\std{0.076} & 0.60 & 5.68\std{0.114} & 19.37 & 20.93\std{0.019} & 16.83 & 22.39\std{0.077} \\
     & Scale-RAE & 19.50 & 26.23\std{0.002} & $-$2.58 & 3.57\std{0.003} & 19.06 & 21.31\std{0.002} & 17.06 & \textbf{25.43}\std{0.014} \\
     & PixArt-Sigma & 21.13 & \textbf{28.07}\std{0.123} & 2.57 & \textbf{9.93}\std{0.117} & 19.38 & \textbf{21.77}\std{0.045} & 15.60 & 23.77\std{0.156} \\
     
    \hline
  \end{tabular}
  }
  \vspace{-0.3cm}
\end{table*}

\begin{table*}[t]
  \centering
  \caption{Per-language image captioning results with BLIP-2 6.7B, Centurio, LLaVA-1.5 7B, and Pangea-7B. The results are reported on the test split. Zero-shot decoding is deterministic (beam search for BLIP-2, greedy for the other models), so a single run is sufficient. Scores of fine-tuned models represent the mean and standard deviation over three training seeds, with the same sampling seed for each image in every run. The average is over the six languages (English is excluded), and its deviation is taken over the three per-seed averages. The best score for each language among both zero-shot and fine-tune models is in bold.}
  \renewcommand{\arraystretch}{1.1}
  \label{tab:captioning_full}
  \vspace{-0.3cm}
  \centering
  \fontsize{7.3}{7.5}\selectfont{
  \begin{tabular}{|l|l |c|c |c|c |c|c |c|c|}
    \hline
     \multirow{2}{*}{Language} & \multirow{2}{*}{Model} & \multicolumn{2}{c|}{spBLEU} & \multicolumn{2}{c|}{chrF++} & \multicolumn{2}{c|}{BERTScore} & \multicolumn{2}{c|}{COMET} \\
    \cline{3-10} 
     &  & Zero-shot & Fine-tuned & Zero-shot & Fine-tuned & Zero-shot & Fine-tuned & Zero-shot & Fine-tuned \\
    \hline
    \multirow{4}{*}{English} & BLIP-2 6.7B & \textbf{12.7} & -- & \textbf{31.9} & -- & 0.771 & -- & 0.681 & -- \\
     & Centurio & 11.5 & -- & 30.8 & -- & \textbf{0.799} & -- & \textbf{0.685} & -- \\
     & LLaVA-1.5 7B & 2.1 & -- & 21.4 & -- & 0.694 & -- & 0.629 & -- \\
     & Pangea-7B & 0.8 & -- & 12.4 & -- & 0.647 & -- & 0.594 & -- \\
    \hline
    \hline
    \multirow{4}{*}{Arabic} & BLIP-2 6.7B & 0.0 & 13.7\std{0.19} & 0.0 & 24.9\std{0.23} & 0.654 & 0.778\std{0.0005} & 0.360 & 0.665\std{0.0022} \\
     & Centurio & 7.8 & 15.1\std{0.17} & 19.4 & \textbf{26.9}\std{0.19} & 0.758 & 0.788\std{0.0006} & 0.635 & \textbf{0.693}\std{0.0019} \\
     & LLaVA-1.5 7B & 0.3 & 13.5\std{0.20} & 9.4 & 24.5\std{0.28} & 0.640 & 0.780\std{0.0016} & 0.378 & 0.670\std{0.0030} \\
     & Pangea-7B & 1.6 & \textbf{15.2}\std{0.17} & 12.8 & 26.5\std{0.20} & 0.659 & \textbf{0.790}\std{0.0007} & 0.644 & 0.687\std{0.0011} \\
     \hline
     \hline
    \multirow{4}{*}{Danish} & BLIP-2 6.7B & 0.2 & 14.5\std{0.09} & 11.5 & 31.7\std{0.17} & 0.699 & 0.784\std{0.0010} & 0.521 & \textbf{0.702}\std{0.0009} \\
     & Centurio & 9.8 & 14.2\std{0.79} & 27.3 & \textbf{32.1}\std{0.49} & 0.776 & 0.787\std{0.0021} & 0.647 & 0.700\std{0.0042} \\
     & LLaVA-1.5 7B & 0.6 & 13.9\std{0.16} & 14.2 & 30.7\std{0.10} & 0.675 & 0.786\std{0.0004} & 0.532 & 0.696\std{0.0030} \\
     & Pangea-7B & 0.5 & \textbf{15.0}\std{0.19} & 12.0 & 31.9\std{0.21} & 0.639 & \textbf{0.790}\std{0.0002} & 0.605 & 0.701\std{0.0017} \\
     \hline
     \hline
    \multirow{4}{*}{French} & BLIP-2 6.7B & 0.1 & 15.6\std{0.08} & 10.9 & 32.1\std{0.14} & 0.715 & 0.798\std{0.0006} & 0.466 & 0.647\std{0.0014} \\
     & Centurio & 11.5 & 15.5\std{0.66} & 28.9 & 32.8\std{0.49} & 0.793 & 0.801\std{0.0016} & 0.612 & \textbf{0.656}\std{0.0047} \\
     & LLaVA-1.5 7B & 2.6 & 14.3\std{0.16} & 23.6 & 31.3\std{0.11} & 0.708 & 0.801\std{0.0013} & 0.586 & 0.641\std{0.0027} \\
     & Pangea-7B & 2.0 & \textbf{16.3}\std{0.19} & 17.8 & \textbf{33.2}\std{0.04} & 0.675 & \textbf{0.805}\std{0.0006} & 0.600 & 0.653\std{0.0011} \\
    \hline
    \hline
    \multirow{4}{*}{Hindi} & BLIP-2 6.7B & 0.0 & 14.0\std{0.18} & 0.0 & 25.9\std{0.10} & 0.648 & 0.786\std{0.0007} & 0.312 & 0.567\std{0.0024} \\
     & Centurio & 7.7 & 12.3\std{0.13} & 20.0 & 25.0\std{0.08} & 0.773 & 0.788\std{0.0003} & 0.501 & 0.577\std{0.0012} \\
     & LLaVA-1.5 7B & 0.3 & 14.2\std{0.38} & 8.6 & 26.2\std{0.20} & 0.620 & 0.791\std{0.0007} & 0.274 & 0.586\std{0.0011} \\
     & Pangea-7B & 2.4 & \textbf{15.7}\std{0.25} & 17.7 & \textbf{28.0}\std{0.15} & 0.675 & \textbf{0.797}\std{0.0004} & 0.550 & \textbf{0.608}\std{0.0005} \\
    \hline
    \hline
    \multirow{4}{*}{Italian} & BLIP-2 6.7B & 0.2 & 15.9\std{0.08} & 12.2 & 32.6\std{0.06} & 0.709 & 0.791\std{0.0005} & 0.471 & \textbf{0.677}\std{0.0012} \\
     & Centurio & 10.5 & 15.0\std{0.48} & 28.0 & 32.5\std{0.34} & 0.785 & 0.793\std{0.0018} & 0.636 & 0.673\std{0.0008} \\
     & LLaVA-1.5 7B & 2.0 & 14.6\std{0.07} & 22.5 & 31.2\std{0.03} & 0.702 & 0.793\std{0.0002} & 0.602 & 0.663\std{0.0014} \\
     & Pangea-7B & 1.7 & \textbf{16.6}\std{0.11} & 17.4 & \textbf{33.3}\std{0.09} & 0.666 & \textbf{0.799}\std{0.0001} & 0.632 & \textbf{0.677}\std{0.0019} \\
    \hline
    \hline
    \multirow{4}{*}{Romanian} & BLIP-2 6.7B & 0.1 & 16.2\std{0.10} & 10.5 & 29.5\std{0.14} & 0.682 & 0.780\std{0.0007} & 0.438 & 0.683\std{0.0011} \\
     & Centurio & 9.4 & 16.0\std{0.17} & 24.2 & 30.1\std{0.15} & 0.763 & 0.785\std{0.0003} & 0.610 & 0.681\std{0.0011} \\
     & LLaVA-1.5 7B & 1.1 & 14.8\std{0.39} & 14.9 & 28.4\std{0.27} & 0.654 & 0.783\std{0.0012} & 0.412 & 0.671\std{0.0018} \\
     & Pangea-7B & 2.0 & \textbf{16.9}\std{0.14} & 16.3 & \textbf{30.4}\std{0.08} & 0.661 & \textbf{0.790}\std{0.0005} & 0.670 & \textbf{0.687}\std{0.0006} \\
    \hline
    \hline
    \multirow{4}{*}{Average} & BLIP-2 6.7B & 0.1 & 15.0\std{0.03} & 7.5 & 29.5\std{0.03} & 0.684 & 0.786\std{0.0001} & 0.428 & 0.657\std{0.0003} \\
     & Centurio & 9.4 & 14.7\std{0.20} & 24.6 & 29.9\std{0.11} & 0.775 & 0.790\std{0.0007} & 0.607 & 0.663\std{0.0011} \\
\textbf{}     & LLaVA-1.5 7B & 1.2 & 14.2\std{0.17} & 15.5 & 28.7\std{0.11} & 0.666 & 0.789\std{0.0008} & 0.464 & 0.655\std{0.0014} \\
     & Pangea-7B & 1.7 & \textbf{15.9}\std{0.04} & 15.6 & \textbf{30.6}\std{0.06} & 0.662 & \textbf{0.795}\std{0.0002} & 0.617 & \textbf{0.669}\std{0.0008} \\
    \hline
  \end{tabular}
  }
  \vspace{-0.15cm}
\end{table*}

\begin{table*}[t]
\centering
\caption{Per-language content-based image retrieval results with CLIP, XLM-R-based CLIP, and BLIP-2. The results are reported on the test split. The averages are computed over six languages (English is excluded). The best score for each language among both zero-shot and fine-tune models is in bold.}
\renewcommand{\arraystretch}{1.1}
\setlength{\tabcolsep}{3pt}
\vspace{-0.3cm}
\fontsize{7.3}{7.5}\selectfont{
\begin{tabular}{|l|l|c|c|c|c|c|c|c|c|}
\hline
\multirow{2}{*}{Language} & \multirow{2}{*}{Model} & \multicolumn{2}{c|}{P@10} & \multicolumn{2}{c|}{R@10} & \multicolumn{2}{c|}{MRR} & \multicolumn{2}{c|}{nDCG@10} \\
\cline{3-4} \cline{5-6} \cline{7-8} \cline{9-10}
 &  & Zero-shot & Fine-tuned & Zero-shot & Fine-tuned & Zero-shot & Fine-tuned & Zero-shot & Fine-tuned \\
\hline
\multirow{3}{*}{English} & CLIP & 0.177 & -- & 0.328 & -- & 0.471 & -- & 0.356 & -- \\
 & XLM-R CLIP & 0.206 & -- & 0.361 & -- & 0.552 & -- & 0.420 & -- \\
 & BLIP-2 & \textbf{0.253} & -- & \textbf{0.395} & -- & \textbf{0.619} & -- & \textbf{0.484} & -- \\
\hline
\hline
\multirow{3}{*}{Arabic} & CLIP & 0.001 & 0.120 & 0.003 & 0.139 & 0.004 & 0.276 & 0.003 & 0.184 \\
 & XLM-R CLIP & 0.130 & \textbf{0.200} & 0.221 & \textbf{0.311} & 0.342 & \textbf{0.492} & 0.245 & \textbf{0.369} \\
 & BLIP-2 & 0.001 & 0.059 & 0.004 & 0.035 & 0.003 & 0.119 & 0.003 & 0.070 \\
\hline
\hline
\multirow{3}{*}{Danish } & CLIP & 0.013 & 0.174 & 0.021 & 0.254 & 0.045 & 0.410 & 0.023 & 0.303 \\
 & XLM-R CLIP & 0.159 & \textbf{0.218} & 0.275 & \textbf{0.356} & 0.429 & \textbf{0.543} & 0.315 & \textbf{0.418} \\
 & BLIP-2 & 0.022 & 0.176 & 0.033 & 0.207 & 0.063 & 0.365 & 0.037 & 0.272 \\
\hline
\hline
\multirow{3}{*}{French } & CLIP &0.063  &0.194  &0.092  &0.299  &0.162  & 0.456 & 0.103 &0.352  \\
 & XLM-R CLIP &0.171  & \textbf{0.217} & 0.260 & \textbf{0.346} &0.400  & \textbf{0.535} & 0.293 & \textbf{0.409}  \\
 & BLIP-2 &0.069  &0.207  &0.112  &0.259  & 0.180 & 0.445 & 0.118 & 0.337 \\

\hline
\hline
\multirow{3}{*}{Hindi} & CLIP & 0.000 & 0.102 & 0.000 & 0.092 & 0.001 & 0.217 & 0.000 & 0.139 \\
 & XLM-R CLIP & 0.119 & \textbf{0.213} & 0.191 & \textbf{0.322} & 0.307 & \textbf{0.500} & 0.218 & \textbf{0.382} \\
  & BLIP-2 & 0.000 & 0.055 & 0.001 & 0.027 & 0.001 & 0.114 & 0.001 & 0.065 \\
\hline
\hline
\multirow{3}{*}{Italian} & CLIP & 0.038 & 0.188 & 0.051 & 0.271 & 0.106 & 0.427 & 0.061 & 0.324 \\
 & XLM-R CLIP & 0.174 & \textbf{0.225} & 0.304 & \textbf{0.368} & 0.457 & \textbf{0.572} & 0.340 & \textbf{0.435} \\
 
 & BLIP-2 & 0.032 & 0.192 & 0.044 & 0.234 & 0.090 & 0.410 & 0.052 & 0.303 \\

\hline
\hline
\multirow{3}{*}{Romanian } & CLIP & 0.015 & 0.173 & 0.015 & 0.251 & 0.044 & 0.395 & 0.022 & 0.295 \\
 & XLM-R CLIP & 0.153 & \textbf{0.218} & 0.269 & \textbf{0.353} & 0.409 & \textbf{0.537} & 0.299 & \textbf{0.413} \\
 & BLIP-2 & 0.017 & 0.177 & 0.023 & 0.215 & 0.049 & 0.372 & 0.026 & 0.274 \\

\hline
\hline
\multirow{3}{*}{Average} & CLIP &0.021  &0.158  &0.030  &0.218  &0.060 &0.356  &0.035  &0.266  \\
 & XLM-R CLIP &0.151  &\textbf{0.217}  &0.260  &\textbf{0.346}  &0.400 &\textbf{0.533}  & 0.293 &\textbf{0.409}  \\

 & BLIP-2 &0.023  &0.144  &0.036  &0.163  &0.064  &0.304  &0.039  &0.220  \\
\hline
\end{tabular}
}
\label{tab:retrieval_metrics_full}
\vspace{-0.3cm}
\end{table*}

\begin{figure*}[t]
  \centering
  \includegraphics[width=\textwidth]{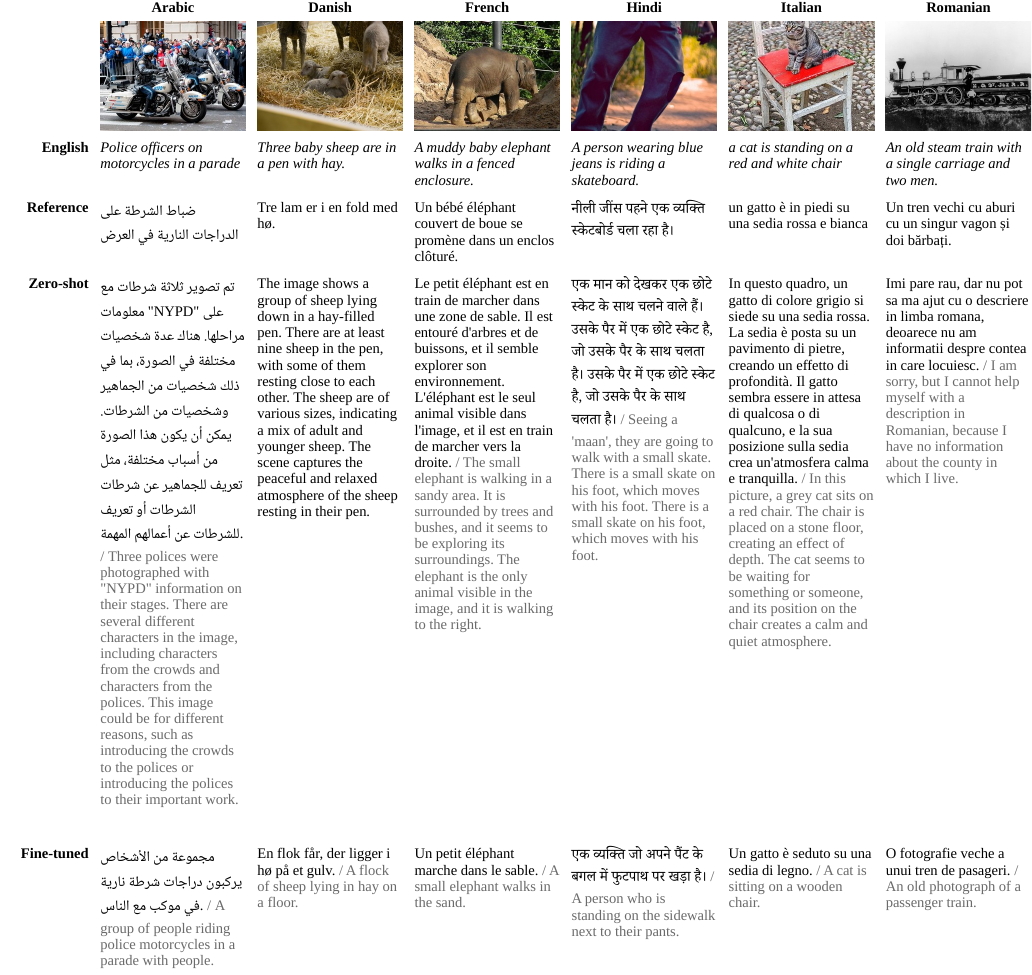}
  \vspace{-0.65cm}
  \caption{Image captioning examples (one target language per image) for LLaVA before and after fine-tuning on \pamuk, with the complete zero-shot outputs. English translations of the captions returned by LLaVA follow the slash (in gray).}
  \label{fig:captioning_examples_full}
  \vspace{-0.3cm}
\end{figure*}

\begin{table*}[t]
\caption{Per-language results for generation and retrieval performance prediction with BERT, Long-CLIP and Correlation CNN. We report Pearson and Kendall correlations with HBPP for GLIDE and SDXL, and with P@10 and RR for CLIP and BLIP-2. Zero-shot denotes English-trained predictors evaluated on the target language, while fine-tuned denotes evaluation after target-language fine-tuning. English results and the original PQPP baselines are included for reference. For each language and predictor, the best of the two settings is shown in bold.}
\renewcommand{\arraystretch}{1.1}
\label{tab:difficulty_full}
\centering
\setlength{\tabcolsep}{2.6pt}
\vspace{-0.3cm}
\fontsize{7.3}{7.5}\selectfont{
\begin{tabular}{@{}|l|l|l|c|c|c|c||c|c|c|c|c|c|c|c|}
\hline
 & & & \multicolumn{4}{c||}{Generative Task}
   & \multicolumn{8}{c|}{Retrieval Task} \\
\cline{4-7}
\cline{8-15}
Language & Predictor & Setting
 & \multicolumn{2}{c|}{HBPP\,/\,GLIDE}
 & \multicolumn{2}{c||}{HBPP\,/\,SDXL}
 & \multicolumn{2}{c|}{P@10\,/\,CLIP}
 & \multicolumn{2}{c|}{RR\,/\,CLIP}
 & \multicolumn{2}{c|}{P@10\,/\,BLIP-2}
 & \multicolumn{2}{c|}{RR\,/\,BLIP-2} \\
\cline{4-5}
\cline{6-7}
\cline{8-9}
\cline{10-11}
\cline{12-13}
\cline{14-15}
& & & Pearson & Kendall
   & Pearson & Kendall
   & Pearson & Kendall
   & Pearson & Kendall
   & Pearson & Kendall
   & Pearson & Kendall \\
\hline
\multirow{6}{*}{English} & \multirow{2}{*}{\begin{tabular}[c]{@{}l@{}} BERT\end{tabular}} & BERT-cased \cite{poesina_CVPR_2025} & 0.566 & 0.406 & 0.281 & 0.232 & 0.451 & 0.277 & 0.221 & 0.176 & 0.511 & 0.328 & 0.168 & 0.139 \\
\textbf{} &  & mBERT (ours) & 0.601 & 0.432 & 0.329 & 0.230 & 0.387 & 0.256 & 0.230 & 0.163 & 0.450 & 0.291 & 0.194 & 0.146 \\
\cline{2-15}
 & \multirow{2}{*}{\begin{tabular}[c]{@{}l@{}}Long-CLIP\end{tabular}} & CLIP \cite{poesina_CVPR_2025} & 0.649 & 0.474 & \textbf{0.380} & \textbf{0.246} & 0.473 & 0.299 & 0.200 & 0.149 & 0.498 & 0.358 & 0.166 & 0.150 \\
 &  & XLM-R CLIP (ours) & \textbf{0.671} & \textbf{0.491} & 0.337 & 0.224 & \textbf{0.599} & \textbf{0.375} & \textbf{0.335} & \textbf{0.235} & \textbf{0.644} & \textbf{0.444} & \textbf{0.281} & \textbf{0.208} \\
\cline{2-15}
\textbf{} & \multirow{2}{*}{\begin{tabular}[c]{@{}l@{}}Corr.  CNN\end{tabular}} & Corr.~CNN \cite{poesina_CVPR_2025} & 0.548 & 0.393 & 0.159 & 0.107 & 0.270 & 0.186 & 0.189 & 0.162 & 0.159 & 0.133 & 0.206 & 0.158 \\
\textbf{} &  & Corr.~CNN (ours) & 0.586 & 0.426 & 0.124 & 0.046 & 0.338 & 0.213 & 0.256 & 0.190 & 0.305 & 0.220 & 0.176 & 0.130 \\
\hline
\hline
\multirow{6}{*}{Arabic} & \multirow{2}{*}{\begin{tabular}[c]{@{}l@{}} BERT\end{tabular}} & Zero-shot & 0.463 & 0.327 & 0.168 & 0.135 & 0.198 & 0.141 & 0.133 & 0.091 & 0.194 & 0.134 & 0.044 & 0.052 \\
 &  & Fine-tuned & 0.562 & 0.404 & 0.276 & 0.203 & 0.344 & 0.218 & 0.222 & 0.169 & 0.394 & 0.262 & \textbf{0.209} & \textbf{0.153} \\
\cline{2-15}
 & \multirow{2}{*}{\begin{tabular}[c]{@{}l@{}}Long-CLIP\end{tabular}} & Zero-shot & 0.665 & \textbf{0.486} & 0.312 & \textbf{0.206} & 0.512 & 0.346 & 0.461 & 0.360 & 0.493 & 0.342 & 0.108 & 0.079 \\
 &  & Fine-tuned & \textbf{0.666} & \textbf{0.486} & \textbf{0.316} & 0.204 & \textbf{0.625} & \textbf{0.403} & \textbf{0.508} & \textbf{0.401} & \textbf{0.645} & \textbf{0.385} & 0.100 & 0.074 \\
\cline{2-15}
 & \multirow{2}{*}{\begin{tabular}[c]{@{}l@{}}Corr.  CNN\end{tabular}} & Zero-shot & 0.586 & 0.426 & 0.124 & 0.046 & 0.088 & 0.074 & 0.049 & 0.052 & 0.305 & 0.290 & 0.087 & 0.069 \\
 &  & Fine-tuned & 0.586 & 0.426 & 0.124 & 0.046 & 0.202 & 0.138 & 0.128 & 0.129 & 0.364 & 0.318 & 0.100 & 0.074 \\
\hline
\hline
\multirow{6}{*}{Danish} & \multirow{2}{*}{\begin{tabular}[c]{@{}l@{}} BERT\end{tabular}} & Zero-shot & 0.520 & 0.371 & 0.206 & 0.143 & 0.271 & 0.181 & 0.170 & 0.124 & 0.275 & 0.205 & 0.099 & 0.085 \\
 &  & Fine-tuned & 0.581 & 0.419 & 0.314 & \textbf{0.230} & 0.369 & 0.226 & 0.214 & 0.156 & 0.446 & 0.304 & \textbf{0.203} & \textbf{0.148} \\
\cline{2-15}
 & \multirow{2}{*}{\begin{tabular}[c]{@{}l@{}}Long-CLIP\end{tabular}} & Zero-shot & \textbf{0.667} & \textbf{0.487} & 0.319 & 0.214 & 0.545 & 0.377 & 0.405 & 0.335 & 0.560 & 0.404 & 0.156 & 0.106 \\
 &  & Fine-tuned & 0.666 & \textbf{0.487} & \textbf{0.324} & 0.211 & \textbf{0.597} & \textbf{0.402} & \textbf{0.439} & \textbf{0.354} & \textbf{0.611} & \textbf{0.434} & 0.171 & 0.115 \\
\cline{2-15}
 & \multirow{2}{*}{\begin{tabular}[c]{@{}l@{}}Corr.  CNN\end{tabular}} & Zero-shot & 0.559 & 0.400 & 0.181 & 0.120 & 0.319 & 0.171 & 0.222 & 0.169 & 0.286 & 0.170 & 0.161 & 0.117 \\
 &  & Fine-tuned & 0.564 & 0.405 & 0.171 & 0.111 & 0.293 & 0.164 & 0.211 & 0.156 & 0.283 & 0.172 & 0.163 & 0.121 \\
\hline
\hline
\multirow{6}{*}{French} & \multirow{2}{*}{\begin{tabular}[c]{@{}l@{}} BERT\end{tabular}} & Zero-shot & 0.539 & 0.385 & 0.236 & 0.164 & 0.236 & 0.167 & 0.217 & 0.143 & 0.289 & 0.193 & 0.130 & 0.101 \\
 &  & Fine-tuned & 0.583 & 0.416 & 0.305 & 0.212 & 0.357 & 0.235 & 0.225 & 0.156 & 0.443 & 0.293 & \textbf{0.204} & \textbf{0.149} \\
\cline{2-15}
 & \multirow{2}{*}{\begin{tabular}[c]{@{}l@{}}Long-CLIP\end{tabular}} & Zero-shot & \textbf{0.668} & \textbf{0.488} & \textbf{0.330} & \textbf{0.219} & 0.557 & 0.359 & \textbf{0.407} & \textbf{0.303} & 0.572 & 0.397 & 0.183 & 0.129 \\
 &  & Fine-tuned & \textbf{0.668} & \textbf{0.488} & 0.329 & 0.217 & \textbf{0.592} & \textbf{0.363} & 0.390 & 0.286 & \textbf{0.613} & \textbf{0.413} & 0.191 & 0.131 \\
\cline{2-15}
 & \multirow{2}{*}{\begin{tabular}[c]{@{}l@{}}Corr.  CNN\end{tabular}} & Zero-shot & 0.559 & 0.400 & 0.181 & 0.120 & 0.322 & 0.170 & 0.191 & 0.146 & 0.279 & 0.170 & 0.150 & 0.108 \\
 &  & Fine-tuned & 0.557 & 0.398 & 0.189 & 0.128 & 0.260 & 0.147 & 0.200 & 0.146 & 0.279 & 0.167 & 0.162 & 0.117 \\
\hline
\hline
\multirow{6}{*}{Hindi} & \multirow{2}{*}{\begin{tabular}[c]{@{}l@{}} BERT\end{tabular}} & Zero-shot & 0.436 & 0.313 & 0.143 & 0.123 & 0.205 & 0.124 & 0.139 & 0.088 & 0.258 & 0.182 & 0.059 & 0.038 \\
 &  & Fine-tuned & 0.567 & 0.405 & 0.303 & \textbf{0.210} & 0.373 & 0.242 & 0.222 & 0.164 & 0.440 & 0.289 & \textbf{0.133} & \textbf{0.098} \\
\cline{2-15}
 & \multirow{2}{*}{\begin{tabular}[c]{@{}l@{}}Long-CLIP\end{tabular}} & Zero-shot & 0.664 & 0.484 & 0.315 & 0.203 & 0.471 & 0.337 & 0.432 & 0.379 & 0.540 & 0.359 & 0.033 & 0.008 \\
 &  & Fine-tuned & \textbf{0.667} & \textbf{0.486} & \textbf{0.321} & 0.204 & \textbf{0.606} & \textbf{0.414} & \textbf{0.520} & \textbf{0.445} & \textbf{0.667} & \textbf{0.395} & 0.109 & 0.069 \\
\cline{2-15}
 & \multirow{2}{*}{\begin{tabular}[c]{@{}l@{}}Corr.  CNN\end{tabular}} & Zero-shot & 0.559 & 0.400 & 0.181 & 0.120 & 0.237 & 0.129 & 0.180 & 0.134 & 0.126 & 0.093 & 0.073 & 0.057 \\
\textbf{} &  & Fine-tuned & 0.545 & 0.393 & 0.193 & 0.109 & 0.183 & 0.122 & 0.180 & 0.132 & 0.160 & 0.108 & 0.084 & 0.056 \\
\hline
\hline
\multirow{6}{*}{Italian} & \multirow{2}{*}{\begin{tabular}[c]{@{}l@{}} BERT\end{tabular}} & Zero-shot & 0.537 & 0.387 & 0.241 & 0.160 & 0.245 & 0.172 & 0.202 & 0.140 & 0.296 & 0.202 & 0.101 & 0.086 \\
 &  & Fine-tuned & 0.576 & 0.417 & 0.325 & \textbf{0.245} & 0.361 & 0.221 & 0.212 & 0.160 & 0.394 & 0.267 & \textbf{0.205} & \textbf{0.150} \\
\cline{2-15}
 & \multirow{2}{*}{\begin{tabular}[c]{@{}l@{}}Long-CLIP\end{tabular}} & Zero-shot & \textbf{0.668} & \textbf{0.488} & 0.329 & 0.217 & 0.562 & 0.374 & 0.419 & 0.319 & 0.582 & 0.411 & 0.177 & 0.111 \\
 &  & Fine-tuned & \textbf{0.668} & \textbf{0.488} & \textbf{0.336} & 0.220 & \textbf{0.617} & \textbf{0.405} & \textbf{0.438} & \textbf{0.336} & \textbf{0.634} & \textbf{0.445} & 0.180 & 0.125 \\
\cline{2-15}
 & \multirow{2}{*}{\begin{tabular}[c]{@{}l@{}}Corr.  CNN\end{tabular}} & Zero-shot & 0.559 & 0.400 & 0.181 & 0.120 & 0.297 & 0.156 & 0.192 & 0.139 & 0.292 & 0.180 & 0.124 & 0.088 \\
 &  & Fine-tuned & 0.561 & 0.403 & 0.179 & 0.114 & 0.282 & 0.156 & 0.207 & 0.144 & 0.292 & 0.186 & 0.154 & 0.112 \\
\hline
\hline
\multirow{6}{*}{Romanian} & \multirow{2}{*}{\begin{tabular}[c]{@{}l@{}} BERT\end{tabular}} & Zero-shot & 0.517 & 0.370 & 0.188 & 0.135 & 0.212 & 0.145 & 0.160 & 0.108 & 0.251 & 0.169 & 0.062 & 0.062 \\
 &  & Fine-tuned & 0.569 & 0.404 & 0.301 & 0.201 & 0.337 & 0.224 & 0.182 & 0.129 & 0.404 & 0.266 & 0.182 & 0.133 \\
\cline{2-15}
 & \multirow{2}{*}{\begin{tabular}[c]{@{}l@{}}Long-CLIP\end{tabular}} & Zero-shot & 0.667 & 0.487 & 0.311 & 0.207 & 0.508 & 0.335 & \textbf{0.423} & \textbf{0.315} & 0.537 & 0.368 & 0.149 & 0.102 \\
 &  & Fine-tuned & \textbf{0.668} & \textbf{0.488} & \textbf{0.321} & \textbf{0.214} & \textbf{0.567} & \textbf{0.359} & 0.408 & 0.312 & \textbf{0.582} & \textbf{0.398} & 0.132 & 0.089 \\
\cline{2-15}
 & \multirow{2}{*}{\begin{tabular}[c]{@{}l@{}}Corr.  CNN\end{tabular}} & Zero-shot & 0.560 & 0.398 & 0.146 & 0.119 & 0.317 & 0.169 & 0.209 & 0.150 & 0.264 & 0.152 & 0.134 & 0.095 \\
 &  & Fine-tuned & 0.586 & 0.426 & 0.124 & 0.046 & 0.312 & 0.212 & 0.257 & 0.196 & 0.322 & 0.239 & \textbf{0.279} & \textbf{0.232} \\
 \hline
 \hline
 \multirow{6}{*}{Average}
 &\multirow{2}{*}{BERT}
 &Zero-shot
 &0.502 &0.359 &0.197 &0.143
 &0.228 &0.155 &0.170 &0.116
 &0.261 &0.181 &0.083 &0.071 \\
&&Fine-tuned
&0.573 &0.411 &0.304 &\textbf{0.217}
 &0.357 &0.228 &0.213 &0.156
 &0.420 &0.280 &\textbf{0.189} &\textbf{0.139} \\
 \cline{2-15}
 & \multirow{2}{*}{Long-CLIP}
 & Zero-shot &\textbf{0.667} &\textbf{0.487} &0.319 &0.211 &0.526 &0.335 &0.425 &0.335 &0.547 &0.380 &0.134 &0.0089 \\
 & & Fine-tuned &\textbf{0.667} &\textbf{0.487} &\textbf{0.325} &0.212 &\textbf{0.601} &\textbf{0.391} &\textbf{0.451} &\textbf{0.356} &\textbf{0.625} &\textbf{0.412} &0.147 &0.101\\
\cline{2-15}

 & \multirow{2}{*}{Corr.~CNN}
 & Zero-shot &0.564 &0.404 &0.166 &0.108 &0.263 &0.145 &0.174 &0.132 &0.259 &0.176 &0.122 &0.089 \\
 & & Fine-tuned &0.567 &0.409 &0.163 &0.092 &0.255 &0.157 &0.197 &0.151 &0.283 &0.198 &0.157 &0.119 \\
\hline
\end{tabular}}
\vspace{-0.3cm}
\end{table*}

\section{Model and Hyperparameter Details}
\label{supp_sec_models}

\subsection{Chosen models}
\label{supp_subsec_models}

\noindent 
\textbf{Text-to-image generation.}
For text-to-image generation, we consider four state-of-the-art generation models that differ in their denoiser and in the way the prompt is encoded. Stable Diffusion 1.5~\cite{rombach2022high} combines a U-Net denoiser with a single CLIP text encoder. Stable Diffusion 3.5 Medium~\cite{esser2024scaling} uses a multimodal diffusion transformer conditioned by two CLIP encoders and a T5 text encoder, while PixArt-Sigma~\cite{chen2024pixart} uses a diffusion transformer conditioned only by a T5 text encoder. These text encoders are mainly pre-trained on English text. In contrast, Scale-RAE~\cite{tong2026scaling} conditions its diffusion transformer on Qwen, a multilingual large language model. The four models thus show whether our data helps, regardless of the denoiser and the text encoder. For Stable Diffusion 1.5, LoRA is applied to both image and text encoders, while for the other three models, LoRA is applied only to the text encoders. These are trained to reproduce, for a prompt in the target language, the output image of the original model for the parallel English prompt. The hyperparameters are tuned by training on the training split and evaluating on the validation split. The final models are then fine-tuned on the COCO prompts of the combined training and validation splits.



\noindent 
\textbf{Image captioning.}
For image captioning, we consider four models that differ in their architecture and exposure to multilingual data. BLIP-2~\cite{Li-ICML-2023} connects a frozen image encoder to a language model through a Q-Former, while LLaVA-1.5~\cite{liu2024improved} connects a CLIP image encoder to a language model through an MLP projector. Both rely on language models mainly pre-trained on English, so they show how far a model with little multilingual training can be adapted using our data alone. In contrast, Centurio~\cite{geigle2025centurio} and Pangea~\cite{yue2025pangea} are multilingual by design, being trained on data covering 100 and 39 languages, respectively, so they show whether our data still helps models that are already multilingual. We caption the test images with each model, in both zero-shot and LoRA fine-tuning settings. We fine-tune models on the training split of each language, with the hyperparameters tuned on the validation split.

\noindent 
\textbf{Content-based image retrieval.}
For content-based image retrieval, we evaluate BLIP-2~\cite{Li-ICML-2023}, the original CLIP based on ViT-B/32~\cite{radford2021learning}, and a modified CLIP based on ViT-B/32, where the text encoder is the multilingual XLM-R model~\cite{conneau-etal-2020-unsupervised}. Each model ranks the images in the MS COCO dataset according to their cosine similarity to the text query. We evaluate each model in zero-shot and fine-tuning scenarios, where the language-specific fine-tuning is applied to the text encoders.

\noindent
\textbf{Prompt/query difficulty prediction.}
For generation and retrieval difficulty prediction, we follow PQPP~\cite{poesina_CVPR_2025} and consider three predictors: BERT, Long-CLIP and Correlation CNN. We replace the English BERT encoder used in PQPP with the multilingual BERT~\cite{Devlin-NAACL-2019}, and fine-tune the encoder and the task-specific regression head. The CLIP-based predictor uses both the prompt and the generated or retrieved images, as it belongs to the category of post-generation/retrieval predictors. We use a frozen XLM-R-based Long-CLIP to support multilingual text, and only train an MLP prediction head. For generation, the head takes concatenated text and image embeddings and predicts the HBPP score. For retrieval, it additionally uses the element-wise product and absolute difference of the embeddings to predict the relevance of each of the top 25 images returned by each retriever (CLIP and BLIP-2). The resulting relevance probabilities are aggregated into expected P@10 and reciprocal rank. The Correlation CNN uses correlation matrices constructed from frozen Long-CLIP-B~\cite{Zhang-ECCV-2024} image embeddings, using the generated images for generation difficulty prediction, and the retrieved images for retrieval difficulty prediction, respectively. The fine-tuning involves adapting both image and text encoders of Long-CLIP-B, using the symmetric CLIP contrastive loss on target-language captions paired with their images from the training split. The images are then re-encoded with the adapted image encoder, and the predictor is retrained on correlation matrices constructed from the updated embeddings.

\vspace{-0.1cm}
\subsection{Hyperparameter configurations}
\label{supp_sec_hyper}
\vspace{-0.1cm}

\noindent
\textbf{Text-to-Image Generation.}
For PixArt-Sigma, Stable Diffusion 3.5 and Scale-RAE, we apply LoRA to the attention and feed-forward layers of the text encoders (the Qwen language model, in the case of Scale-RAE), using AdamW with a cosine learning rate schedule and linear warm-up. For PixArt-Sigma and Stable Diffusion 3.5, the loss is the mean squared error between the output of the denoiser for the prompt in the target language, and the output of the original model for the English prompt, computed on the same noisy latent and timestep. Scale-RAE has no separate text encoder: its language model produces the conditioning sequence received by the diffusion transformer. We therefore apply the mean squared error directly to this conditioning, and add a Kullback-Leibler term (weighted by $0.1$) between the next-token distributions of the two models on the tokens that precede the image. One in four training steps uses the English prompt as input, to preserve the behavior on English. Following the default sampling setup of each model, we generate images with PixArt-Sigma (20 steps, guidance scale $4.5$, $512 \times 512$ pixels), Stable Diffusion 3.5 (40 steps, guidance scale $4.5$, $512 \times 512$ pixels) and Scale-RAE (50 steps, no classifier-free guidance, $224 \times 224$ pixels). For Stable Diffusion 1.5 we fine-tune the attention and feed-forward layers of the CLIP text encoder as well as the attention layer of the U-Net with LoRA (rank $r=16$). We use AdamW with cosine learning rate schedule and 300 warm-up steps. Training is performed with batch size 16 for 6000 steps. Of the 6000 steps, the first 1500 only train the text encoder, the rest of the network being frozen. After that, the text encoder and the U-Net are trained jointly. All configurations are selected by HPSv2.1 on the validation split, and are listed in \cref{tab:hyperparameters}.

\noindent
\textbf{Image Captioning.}
For BLIP-2, Centurio, LLaVA-1.5 and Pangea-7B, we apply LoRA with a dropout of $0.05$ to the linear layers of the language model, keeping all the other components frozen, and using AdamW with a cosine learning rate schedule and linear warm-up. BLIP-2 receives only the image as input, while the other three models also receive the instruction ``Briefly describe the image in \{language\}.''. Following the default decoding setup of each model, BLIP-2 uses beam search with 5 beams, while the other three models use greedy decoding. Configurations for all languages are selected by the average chrF++ on the Romanian, Hindi and French validation splits, and are listed in \cref{tab:hyperparameters}.

\noindent
\textbf{Generation/difficulty prediction.}
We search for the learning rate $\eta \in \{10^{-5}, 5 \times 10^{-5}, 10^{-4}\}$ and weight decay values $\lambda \in \{0, 0.01, 0.1\}$. Hyperparameters are tuned separately for each language, and the configuration and checkpoint with the lowest validation loss is selected. For the CLIP-based retrieval predictor, selection is based on the Pearson correlation for P@10 on the validation split.

We train the mBERT-based predictor with AdamW for 25 epochs, using a batch size of 256 and dropout of $0.3$. For generation, we train the Long-CLIP predictor with Adam ans MSE loss for 100 epochs. For retrieval, we use AdamW and binary cross-entropy to predict per-image relevance, training for 25 epochs. Both settings use a batch size of 256 and a dropout of $0.5$. We train the Correlation CNN with Adam and MSE loss for 25 epochs, using dropout of $0.5$ and batch size of 128 for generation and 32 for retrieval.

\vspace{-0.1cm}
\section{Detailed Results}
\label{supp_sec_results}
\vspace{-0.1cm}

The aggregated results for all languages are discussed in \cref{sec:experiments}. Here, we discuss the variations across languages and metrics.

\noindent
\textbf{Text-to-image generation.}
The results for text-to-image generation are shown in \cref{tab:t2i_full}. One thing to notice is that the best performers and the effects of fine-tuning tend to be consistent across all metrics for most languages. PixArt-Sigma performs best on all metrics, except on CLIPScore. Scale-RAE generally performs better when measured using CLIPScore, although surprisingly, SD 1.5 wins the zero-shot comparison when measured using this metric. Models perform poorly on Hindi and Arabic when used in a zero-shot manner, while the same models perform better on Italian and French, probably due to the compatible alphabet and their closer similarity to English. Performance levels on Danish and Romanian sit somewhere in the middle. Zero-shot results are comparatively higher for Italian and French, but they still lag behind English. Fine-tuning closes the observed gaps. The effect of fine-tuning is even more pronounced for Arabic and Hindi. This closing gap behavior is consistent across all metrics.

\noindent
\textbf{Image captioning.} 
The results for image captioning are presented in \cref{tab:captioning_full}. The fine-tuned Pangea-7B model generally attains the best performance in all metrics, for all languages, but, unexpectedly, it performs poorly in the zero-shot setting. Hence, due to fine-tuning, it outperforms the other models in most languages and metrics. Most models perform poorly in the zero-shot setting in terms spBLEU, but fine-tuning makes nearly all of them outperform the best zero-shot setting. On Italian, French and Danish, Centurio performs well, relatively close to English, but features a massive drop in performance on Hindi and Arabic. Its performance on Romanian is somewhere in between. The zero-shot Pangea-7B performs better in terms of chrF++ on most languages (except Danish) than on English. The other metrics show a similar behavior, but the effect is less pronounced. All BERTScore values are relatively similar, around a value of $0.7$, but we can observe an improvement after fine-tuning. In general, COMET scores are consistent with the BERTScore values. 

In \cref{fig:captioning_examples_full}, we show the image and the associated texts for six image captioning examples. English translations for all texts are provided for reference. The zero-shot setting sometimes provides answers in English (\eg instead of Danish), complains about not knowing the language (\eg Romanian) or responds with barely comprehensible text (\eg Hindi). Fine-tuning fixes these problems, but the captions are not always perfect (\eg Hindi).

\noindent
\textbf{Content-based image retrieval.}
The results for the content-based image retrieval task are presented in \cref{tab:retrieval_metrics_full}. The four metrics present a remarkably similar picture. BLIP-2 wins the zero-shot English. XLM-R CLIP clearly leads in both zero-shot and fine-tuned settings, on all target languages, but English. It is worth noticing that, for CLIP and BLIP-2, while the performance is low for non-English across the board, the performance on Hindi and Arabic is zero or almost zero. In both cases, fine-tuning helps the models close the gap. XLM-R CLIP performs the best for all languages, and matches its own zero-shot performance on English. However, it still fails to match the performance that BLIP-2 exhibits, when the latter is presented with English prompts.

\noindent
\textbf{Generation difficulty prediction.}
The generation difficulty prediction results are reported in the first part (left-hand side) of \cref{tab:difficulty_full}. In the zero-sot setting, BERT is performing worse on all the metrics and all non-English languages, compared with its performance on English.  Fine-tuning helps, but, although it sometimes closes the gap on one metric for one language, like Danish for the Kendall correlation with HBPP on SDXL, it remains behind when we look at the bigger picture. For all models, performance on Hindi and Arabic is behind the other non-English languages, whereas performance on French is usually better.
The Correlation CNN is largely unaffected by both language and fine-tuning, with fine-tuning decreasing the performance just as often as increasing it, both of those in a marginal and language independent way. Long-CLIP performs close to English on all languages, including Arabic and Hindi. Fine-tuning Long-CLIP mostly leads to marginal improvements, but it sometimes hurts performance, \eg for French. Long-CLIP performs best across the board, with a few exceptions. 

\noindent
\textbf{Retrieval difficulty prediction.}
The results for retrieval difficulty prediction are reported in the second part (right-hand side) of \cref{tab:difficulty_full}. The Correlation CNN often performs worse on the non-English prompts, and fine-tuning helps, but in a consistent way. The performance is lower on Hindi and Arabic. Fine-tuning almost closes the gap for Romanian and Italian, but not for French and Danish. Although the zero-shot BERT performs poorly, the fine-tuned BERT performs surprisingly well, even for Arabic and Hindi. The zero-shot Long-CLIP performs well across languages, for most metrics and tasks, and fine-tuning is usually beneficial. The fine-tuned Long-CLIP sometimes outperforms English and, on average, it closes the gap that exists between English and the other languages.

\end{document}